\documentclass[11pt]{article}

\usepackage[final]{acl}

\usepackage{times}
\usepackage{latexsym}

\usepackage[T1]{fontenc}

\usepackage[utf8]{inputenc}

\usepackage{microtype}

\usepackage{inconsolata}
\usepackage{algorithm}

\usepackage{graphicx}
\usepackage{multirow}
\usepackage{booktabs}
\usepackage{tabularx}
\usepackage{adjustbox}
\usepackage{array}
\usepackage{nccmath}

\usepackage{helvet}  
\usepackage{courier}  

\usepackage{natbib}  
\usepackage{caption} 
\usepackage{algorithmic}
\usepackage{ragged2e}
\usepackage{makecell}   

\usepackage{amsmath,amssymb,amsfonts}
\usepackage{textcomp}
\usepackage{xcolor}
\usepackage{newfloat}
\usepackage{listings}
\usepackage[most]{tcolorbox} 
\definecolor{codegray}{rgb}{0.9,0.9,0.9}
\definecolor{codepink}{rgb}{0.98,0.3,0.5}

\lstdefinestyle{mystyle}{
    backgroundcolor=\color{codegray},
    keywordstyle=\color{green},
    basicstyle=\ttfamily\small,
    breaklines=true,
    language=Verilog,
    frame=tb,
    framerule=0pt,
    xleftmargin=1em,
    numbers=none
}
\title{ChipSeek: Optimizing Verilog Generation via EDA-Integrated Reinforcement Learning}

\author{
 \textbf{Zhirong Chen\textsuperscript{1,2}},
 \textbf{Kaiyan Chang\textsuperscript{1,2}},
 \textbf{Zhuolin Li\textsuperscript{3}},
 \textbf{Cangyuan Li\textsuperscript{1,2}},
 \\
 \textbf{Xinyang He\textsuperscript{4}},
 \textbf{Chujie Chen\textsuperscript{1,2}},
 \textbf{Mengdi Wang\textsuperscript{1,2}},
 \textbf{Haobo Xu\textsuperscript{1,2}},
 \\
 \textbf{Yinhe Han\textsuperscript{1,2}},
 \textbf{Huawei Li\textsuperscript{1,2}},
 \textbf{Ying Wang\textsuperscript{1,2}}
\\
 \textsuperscript{1}SKLP, Institute of Computing Technology, Chinese Academy of Sciences,
\\
 \textsuperscript{2}University of Chinese Academy of Sciences,
\\
 \textsuperscript{3}University of Electronic Science and Technology of China,
\\
\textsuperscript{4}Beijing Institute of Technology
\\
 \small{
   \textbf{Correspondence:} \href{mailto:wangying2009@ict.ac.cn}{wangying2009@ict.ac.cn}
 }
}

\begin{document}
\maketitle
\begin{abstract}
Large Language Models have emerged as powerful tools for automating Register-Transfer Level (RTL) code generation, yet they face critical limitations: existing approaches typically fail to simultaneously optimize functional correctness and hardware efficiency metrics such as Power, Performance, and Area (PPA). Methods relying on supervised fine-tuning commonly produce functionally correct but suboptimal designs due to the lack of inherent mechanisms for learning hardware optimization principles. Conversely, external post-processing techniques aiming to refine PPA performance after generation often suffer from inefficiency and do not improve the LLMs' intrinsic capabilities.

To overcome these challenges, we propose ChipSeek, a novel hierarchical reward based reinforcement learning framework designed to encourage LLMs to generate RTL code that is both functionally correct and optimized for PPA metrics. Our approach integrates direct feedback from EDA simulators and synthesis tools into a hierarchical reward mechanism, facilitating a nuanced understanding of hardware design trade-offs. Through Curriculum-Guided Dynamic Policy Optimization (CDPO), ChipSeek enhances the LLM's ability to generate high-quality, optimized RTL code. Evaluations on standard benchmarks demonstrate ChipSeek’s superior performance, achieving state-of-the-art functional correctness and PPA performance. Furthermore, it excels in specific optimization tasks, consistently yielding highly efficient designs when individually targeting fine-grained optimization goals such as power, delay, and area. The artifact is open-source in \href{https://github.com/rong-hash/chipseek}{https://github.com/rong-hash/chipseek}.
\end{abstract}

\section{Introduction}

Large Language Models (LLMs) show immense potential to revolutionize hardware design methodology, particularly in tasks like Register-Transfer Level (RTL) code generation \cite{10.1145/3676536.3676830, 10.5555/3692070.3693698}. Previous works have improved RTL generation by employing various techniques, including Supervised Fine-tuning (SFT) \cite{thakur2023verigenlargelanguagemodel, Chang_2024}, Retrieval-Augmented Generation (RAG) \cite{10817982}, multi-agent collaboration \cite{yu2025spec2rtlagentautomatedhardwarecode, Ho_Ren_Khailany_2025}, and Chain of Thought (CoT) reasoning \cite{qin2025reasoningvefficientverilogcode}. While these approaches successfully enhance functional correctness, they generally neglect critical hardware metrics such as synthesizability and more importantly, Power, Performance, and Area (PPA).

\begin{figure}[!t] 
    \centering
    \includegraphics[width=\linewidth]{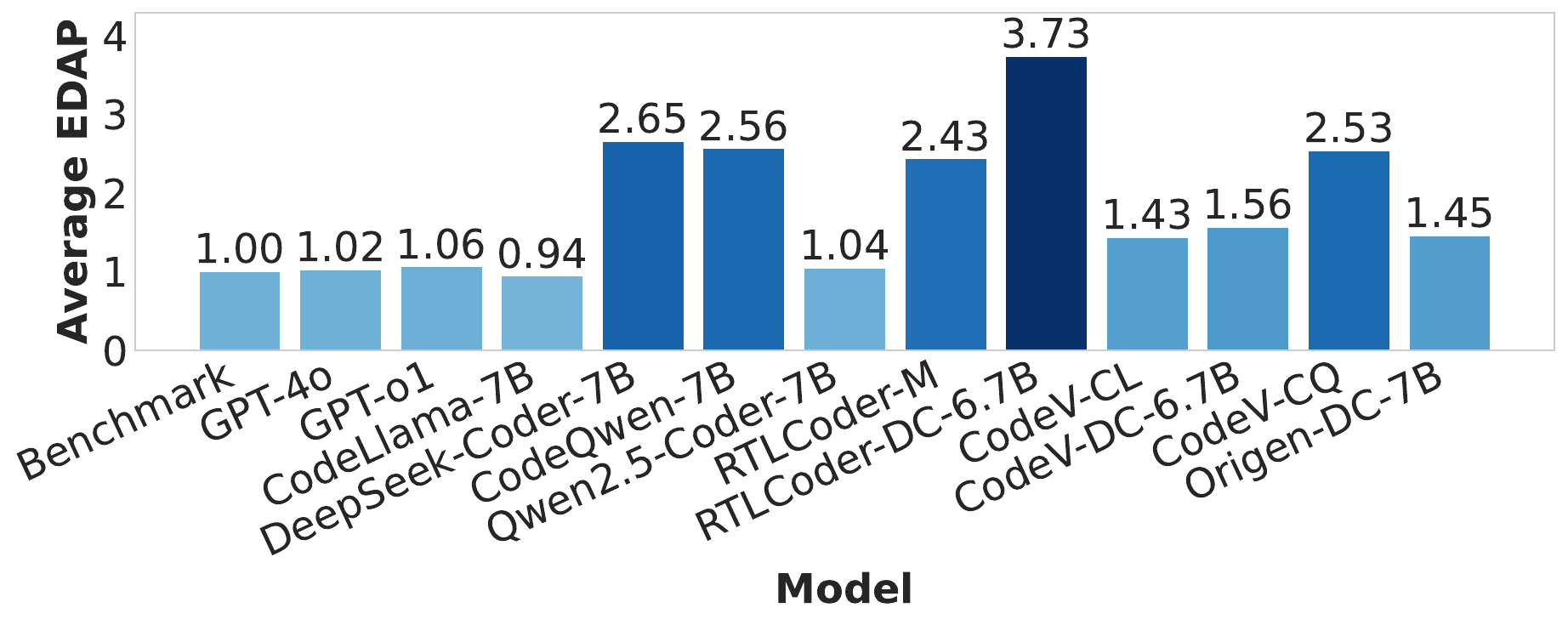}
    \caption{PPA performance comparison between Verilog from models and benchmark RTLLM using the EDAP (Energy-Delay-Area Product). Lower values are better.
}
    \label{fig:visualize_edap}
\end{figure}
However, to truly assist engineers, it is not enough to produce merely functionally correct Verilog. The generated RTL must also be high-quality in terms of PPA metrics. PPA performance is a crucial indicator of Verilog code quality. In practice, design specifications vary widely: edge devices may emphasize area and power constraints, while data center accelerators focus on timing and performance. Existing models, trained predominantly on inconsistent and noisy RTL corpora, lack the inductive bias needed to internalize such nuanced trade-offs, sometimes outcompeted by designs written by expert engineers. 
Figure~\ref{fig:visualize_edap} illustrates this problem clearly: the Verilog generated by existing models mostly underperforms compared to the RTLLM benchmark, a collection of expert-written designs. These models fail to match the hardware efficiency of manually crafted RTL and often require post-processing, such as Monte Carlo Tree Search (MCTS) \cite{delorenzo2024makecountllmbasedhighquality} or external optimization pipelines \cite{yao2024rtlrewritermethodologieslargemodels}, to improve PPA. However, such methods operate externally, introducing computational and manual overhead  to take effect without improving the LLM intrinsic ability. Therefore, there is a fundamental gap: current methods lack an inherent mechanism to optimize functional correctness and hardware-specific PPA metrics concurrently. Bridging this gap is essential for making LLMs viable co-designers in practical RTL workflows.

To address this challenge, we introduce \textbf{ChipSeek}, a novel reinforcement learning framework that integrates EDA toolchains directly into the training loop. The EDA toolchain offers functional verification to ensure logical correctness and PPA metric measurement to quantify hardware efficiency. Together, these feedback signals serve as rewards to enforce both functional validity and alignment with design specifications. Our framework employs a hierarchical reward system and Curriculum-Guided Dynamic Policy Optimization (CDPO), enabling the LLM to adapt the optimization goals with the training process and optimize RTL code generation according to specific PPA targets. This empowers the LLM to generate RTL codes that not only meet functional correctness requirements but also exhibit high quality in terms of hardware PPA. Our main contributions are as follows:

\textbf{Hierarchical Rewards from Comprehensive EDA Toolchain:}
We build a closed-loop RTL generation pipeline that tightly integrates a comprehensive open-source EDA toolchain (compilation, simulation, synthesis, and backend analysis) into reinforcement learning. Based on this toolchain, we derive hierarchical reward signals spanning thinking format, syntax validity, functional correctness, synthesizability, and PPA, together with a strict hierarchical gating mechanism that avoids expensive downstream evaluation on invalid designs. This enables direct tool-verified supervision for functional correctness and provides PPA-aware feedback grounded in physical implementation.

\textbf{Curriculum-Guided Dynamic Policy Optimization (CDPO):}
We propose CDPO for multi-objective RTL optimization under two key challenges in managing EDA-derived rewards: \emph{(i) learning-stage mismatch}, where process rewards such as format reward and syntax reward are easier to learn and serve as early-stage scaffolding while functional correctness and PPA are the ultimate goals; and \emph{(ii) scale mismatch}, arises from the disparity between binary discrete signals (0/1) for syntax or function rewards versus continuous, theoretically unbounded PPA rewards, which renders traditional reward aggregation unstable. CDPO addresses these issues with a curriculum-guided weight schedule, advantage-level aggregation, and prompt-conditioned PPA preference weighting, enabling curriculum training and controllable power--performance--area trade-offs.

\textbf{Automated Data Augmentation Pipeline:}
We propose a multi-stage automated data augmentation pipeline that systematically processes  Verilog codes into richer PPA-aware datasets. This pipeline executes three fundamental tasks: (1) generating reasoning cold-start datasets tailored explicitly for the SFT phase; (2) synthesizing diverse PPA preference vectors and augmenting design descriptions to facilitate comprehensive multi-objective RTL optimization; and (3) producing accurate testbenches and corresponding PPA metrics critical for precise reward computation. Extensive validation and rigorous filtering procedures ensure dataset quality, significantly boosting the robustness and effectiveness of subsequent training processes.

\begin{figure*}[t] 
    \centering
    \includegraphics[width=\textwidth]{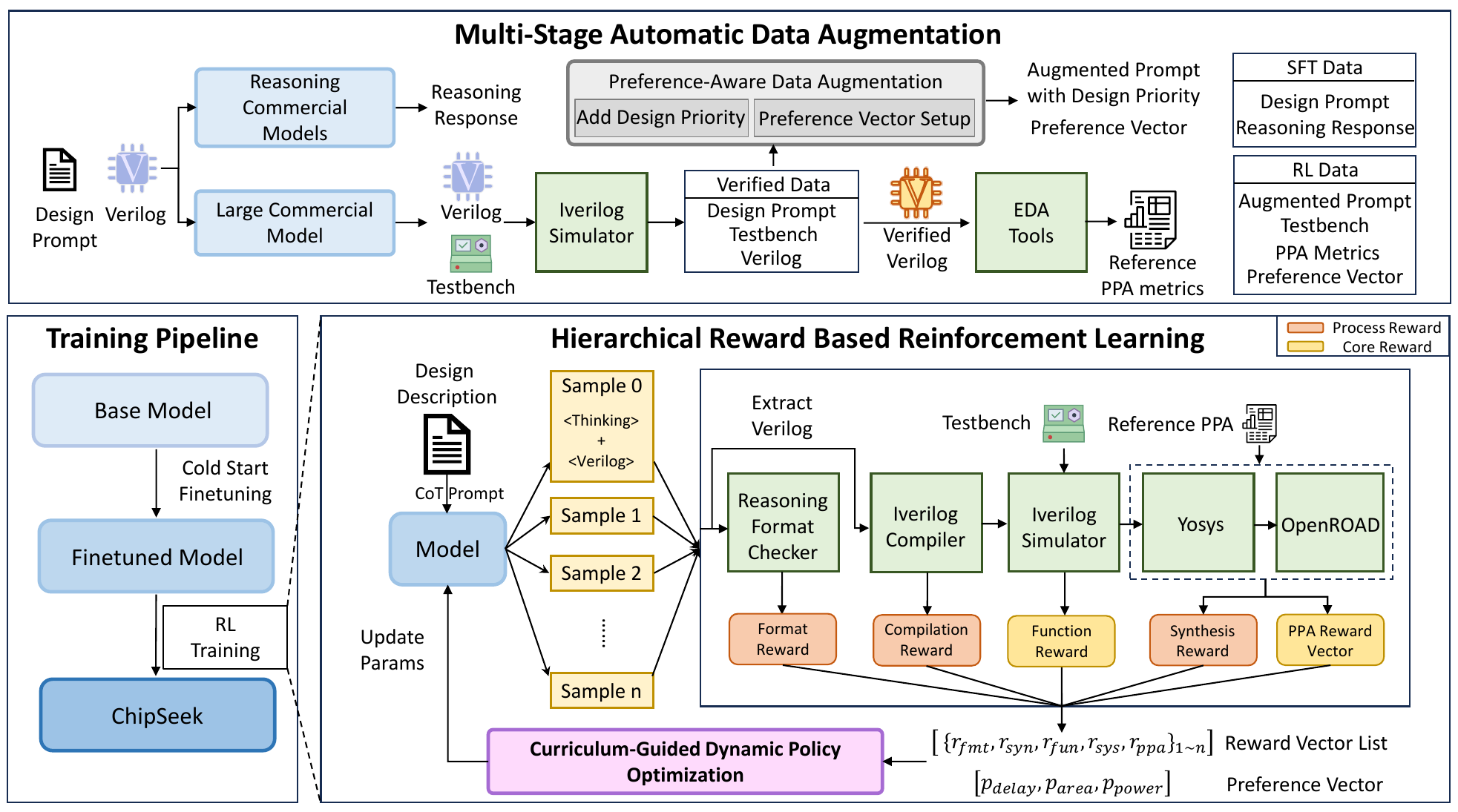}
    \caption{Our Hierarchical Reward based Reinforcement Learning Framework.}
    \label{fig:framework}
\end{figure*}

\section{Background }
\sloppy

Large Language Models (LLMs) have demonstrated significant potential in generating Verilog code directly from natural language specifications. Despite this potential, the quality of the generated Verilog code remains limited due to two fundamental challenges: ensuring functional correctness and achieving optimization in terms of PPA.

Several approaches have been proposed to improve functional correctness in Verilog code generation. For instance, RTLFixer \cite{10.1145/3649329.3657353} and HDLDebugger \cite{yao2024hdldebuggerstreamlininghdldebugging} utilize Retrieval-Augmented Generation (RAG) to facilitate autonomous debugging by LLMs. Additionally, fine-tuning \cite{Chang_2024} and multimodal techniques \cite{10.1145/3676536.3676679} have been explored to enhance functional accuracy. Regarding performance optimization, RTLRewriter and SymRTLO \cite{yao2024rtlrewritermethodologieslargemodels, wang2025symrtloenhancingrtlcode} applies code analysis combined with RAG to pinpoint redundant structures and potential optimizations. VeriGen-MCTS \cite{delorenzo2024makecountllmbasedhighquality} leverages MCTS to optimize hardware performance metrics.

\section{Method}
\label{sec:method}

\subsection{Overview}
Our framework, ChipSeek, aims to align LLMs with the rigorous constraints of hardware design. As illustrated in Figure~\ref{fig:framework}, the framework operates in a closed loop where the LLM acts as a policy $\pi_\theta$, generating Verilog code $y$ given a design specification $x$. The generated code is evaluated by a comprehensive EDA toolchain, providing feedback signals ranging from basic syntax compliance to complex PPA metrics.

To effectively navigate this complex optimization landscape, we propose \textbf{Curriculum-Guided Dynamic Policy Optimization}. Unlike traditional RLVR methods that aggregate rewards into a scalar before optimization, CDPO disentangles distinct feedback signals, normalizes and aggregates them in the advantage space, and dynamically modulates their influence via a curriculum schedule. This schedule guides the model to first master reasoning structures and syntactic compliance, laying the groundwork for subsequent optimization of functional correctness and PPA performance.

\subsection{Hierarchical Rewards Definition}
\label{sec:rewards}
We partition the rewards into \textit{Process Rewards} (Format, Syntax, Synthesis) and \textit{Core Rewards} (functional correctness and PPA). To optimize computational resources, we implement a strict gating mechanism where downstream metrics are only evaluated if upstream constraints are satisfied.

\noindent\textbf{Process Rewards ($\mathcal{R}_{pro}$):} These serve as binary prerequisites for a valid Verilog code. 

\textbf{Format Reward ($r_{fmt}$)} enforces the CoT structure \texttt{(think)...(answer)...} to promote reasoning stability. 

\textbf{Syntax Reward ($r_{syn}$)} is verified by the Icarus Verilog compiler, where $r_{syn}=1$ indicates successful compilation. 

\textbf{Synthesis Reward ($r_{sys}$)} is verified by Yosys and OpenROAD \cite{openroad}, where $r_{sys}=1$ confirms the design is physically synthesizable into a netlist, ensuring hardware realizability beyond logical syntax.

\noindent\textbf{Core Rewards ($\mathcal{R}_{cor}$):} These evaluate the semantic quality and performance of the design. 

\textbf{Function Reward ($r_{func}$)} is determined by testbench simulation, where $r_{func}=1$ only if all test cases pass. 

\textbf{PPA Reward Vector($\mathbf{r}_{ppa}$)} quantifies Power, Delay, and Area rewards. For each metric $m \in \{power, delay, area\}$, the reward is the relative improvement over a reference design: $r_{m} = \text{ref}_m / \text{gen}_m$, where $\text{ref}_m$ and $\text{gen}_m$ are the metric values of the reference design and the generated design.

\noindent\textbf{Gating Mechanism:} 
We impose a hierarchical dependency chain to avoid expensive simulations on invalid code: \textbf{Syntax $\to$ Function $\to$ Synthesis $\to$ PPA}. Specifically, a higher-level reward is computed only if the immediate lower-level reward is positive (e.g., $r_{func}$ is triggered only if $r_{syn}=1$). If a stage fails, the evaluation terminates, and all subsequent rewards are assigned a value of 0.

\subsection{Curriculum-Guided Dynamic Policy Optimization (CDPO)}
\label{sec:cg_dpo}

Given the hierarchical rewards in Section~\ref{sec:rewards}, we optimize the policy with two design choices: (i) \emph{decoupled advantage estimation} for each reward, and (ii) \emph{dynamic policy optimization} that apply self-adaptive weights of rewards over training steps and conditions PPA optimization on prompt-level preferences. Finally, we aggregate all objectives at the \emph{advantage level} and update the policy with the corresponding loss.

\subsubsection{Decoupled Advantage Estimation}
\label{sec:decoupled_adv}

For each prompt $q$, we sample a group of $G$ completions $\{o_i\}_{i=1}^{G}$. Each completion yields a multi-objective reward vector $\mathbf{r}_i$ containing component metrics $r_{k,i}$, where each metric corresponds to the process ($\mathcal{R}_{pro}$) and objective ($\mathcal{R}_{cor}$) definitions detailed in Section~\ref{sec:rewards}.

For a specific component $k$, the decoupled token-level advantage is defined as:
\begin{equation}
\small
\hat{A}^{(k)}_{i,t}=\frac{r_{k,i}-\mu_k}{\sigma_k+\epsilon},\qquad \forall t\in\{1,\dots,|o_i|\},
\label{eq:decoupled_group_adv}
\end{equation}
where $\mu_k$ and $\sigma_k$ denote the mean and standard deviation of the $k$-th reward component across group $G$. This results in a set of advantages that are subsequently aggregated in the dynamic policy optimization phase. Such normalization brings reward components of different scales onto a common scale before aggregation, effectively mitigating cross-component interference caused by reward scale mismatch.

\subsubsection{Dynamic Policy Optimization} 
\label{sec:dynamic_po}

We dynamically weight Process rewards via curriculum annealing, and weight Objective rewards via fixed function reward weight plus preference-conditioned PPA reward weights. All objectives are combined \emph{after} advantage estimation.

\paragraph{Adaptive Curriculum for Process Rewards.}
We utilize a curriculum weight schedule mechanism to stabilize the optimization trajectory. This ensures that the policy initially focuses on easier process rewards (e.g., syntax and format) and gradually shifts its emphasis toward the core objectives (e.g., functional correctness and PPA) as it masters basic Verilog programming patterns, enabling an easy-to-hard learning progression. For each process objective $k \in \mathcal{R}_{pro}$ at training step $s$, we compute the global success rate $\bar{\mu}_k^{(s)}$ over the entire training batch of $B$ prompts and $G$ completions:
\begin{equation}
\small
\bar{\mu}_k^{(s)} = \frac{1}{B \times G} \sum_{j=1}^{B} \sum_{i=1}^{G} r_{k,ji}^{(s)},
\end{equation}
where $r_{k,ji}^{(s)}$ is the reward for the $i$-th completion of the $j$-th prompt. We first compute an instantaneous curriculum signal
\begin{equation}
\small
\hat{\alpha}_k^{(s)} = \max\!\left(0,\, 1 - \bar{\mu}_k^{(s)}\right),
\end{equation}
and then apply an exponential moving average to obtain a smooth curriculum coefficient:
\begin{equation}
\small
\alpha_k^{(s)} = \beta\,\alpha_k^{(s-1)} + (1-\beta)\,\hat{\alpha}_k^{(s)}.
\end{equation}
This ensures the gradient contribution of each process reward changes smoothly across steps, reducing batch-to-batch oscillation. The aggregated process advantage is:
\begin{equation}
\small
A_{i,t}^{pro} = \sum_{k\in \mathcal{R}_{pro}} \alpha_k^{(s)} \hat{A}^{(k)}_{i,t}.
\label{eq:adaptive_process_adv}
\end{equation}

\paragraph{Preference-conditioned Weighting for PPA Rewards.}
We maintain a fixed weight for functional correctness while steering PPA optimization via prompt-dependent preference templates. This design allows the policy to prioritize the corresponding PPA component according to the optimization preference specified in the prompt. Let $\mathbf{p}(q)=(p_P, p_D, p_A)$ denote the preference vector extracted from prompt $q$ (e.g., via tags like \texttt{[low\_power]}), where $\sum p_m=1$. For the $i$-th completion, the aggregated core advantage is:
\begin{equation}
\label{eq:scalar-reward}
\small
A_{i,t}^{cor} = w_{func}\hat{A}^{(func)}_{i,t} + \sum_{m\in\{P,D,A\}} p_m(q)\hat{A}^{(m)}_{i,t}.
\end{equation}

The final token-level advantage for optimization is formed by summing the adaptive process component and the preference-weighted core component:
\begin{equation}
\small
A_{i,t}^{total} = A_{i,t}^{pro} + A_{i,t}^{cor}.
\end{equation}

For each prompt $q$, we sample a group of $G$ rollouts $\{o_i\}_{i=1}^{G}$ from the behavior policy $\pi_{\theta_{\text{old}}}$. We maximize the expected objective:
\begin{equation}
\small
\label{eq:CDPO-objective}
\mathcal{J}(\theta) = \mathbb{E}_{\{o_i\}_{i=1}^G\sim \pi_{\theta_\text{old}}} \Bigg[
\frac{1}{\sum_{i=1}^{G}|o_i|}
\sum_{i=1}^{G}\sum_{t=1}^{|o_i|}
\mathcal{O}_{i,t}(\theta)
\Bigg],
\end{equation}
where the token-level objective $\mathcal{O}_{i,t}(\theta)$ prevents excessive policy updates via decoupled clipping:
\begin{equation}
\label{eq:token-objective}
\small
\begin{aligned}
\mathcal{O}_{i,t}(\theta) = & \min \Big( 
r_{i,t}(\theta) A_{i,t}^{total}, 
\\ & \text{clip}\big(r_{i,t}(\theta), 1-\varepsilon_{\text{low}}, 1+\varepsilon_{\text{high}}\big) A_{i,t}^{total} 
\Big).
\end{aligned}
\end{equation}
The probability ratio $r_{i,t}(\theta)$ is defined as in Eq.~\eqref{eq:advantage_calculation}.
\begin{equation}
    \small
    r_{i,t}(\theta)=\frac{\pi_{\theta}(o_{i,t} \mid q, o_{i,<t})}{\pi_{\theta_{\text{old}}}(o_{i,t} \mid q,o_{i,<t})}.
\label{eq:advantage_calculation}
\end{equation}

\subsection{Multi-Stage Data Generation Framework }
\label{sec:data_augmentation}

As depicted on the upper half of Figure \ref{fig:framework}, our multi-stage automated data augmentation framework comprises three main stages. The first creates a dataset for supervised fine-tuning using CoT reasoning. The subsequent two stages generate data tailored for reinforcement learning stage.

For the initial supervised training, we curate  Verilog code from online repositories, perform syntax checking, and then use the DeepSeek-R1 model to enrich the samples with natural language descriptions and CoT reasoning chains. This foundational dataset fosters the model's initial reasoning and Verilog generation capabilities.

To generate data for accurate rewards computation, we first address the \textbf{functional reward}. We use GPT-5 to generate multi-case testbenches for our Verilog codes. A verification pipeline then ensures that only functionally correct code-testbench pairs are included in the RL training set, providing significant validation feedback.

Next, to provide the \textbf{PPA performance reward}, we use an automated backend pipeline to extract power, area, and delay metrics via NanGate45 process simulations. Validated designs with their PPA metrics are integrated into the dataset.

To enable multi-objective optimization, we employ a specialized data augmentation strategy consisting of two key steps: prompt augmentation and preference vector generation.

For \textbf{Prompt Augmentation}, we augment each instruction in our RL dataset with a fine-grained design priority, such as ``Focus on minimizing the hardware area." This guides the model to optimize for specific PPA metrics.

To complement these prompts, we assign a corresponding \textbf{Preference Vector} $\vec{p}(q)=(p_P, p_D, p_A)$ according to a predefined set of templates. This vector is used to compute the preference-weighted advantage in Eq.~\ref{eq:scalar-reward}. Under the constraint $\sum_{m\in{P,D,A}} p_m=1$, the template ensures the preferred metric receives the largest weight.

\begin{figure}[!t] 
    \centering
    \includegraphics[width=\linewidth]{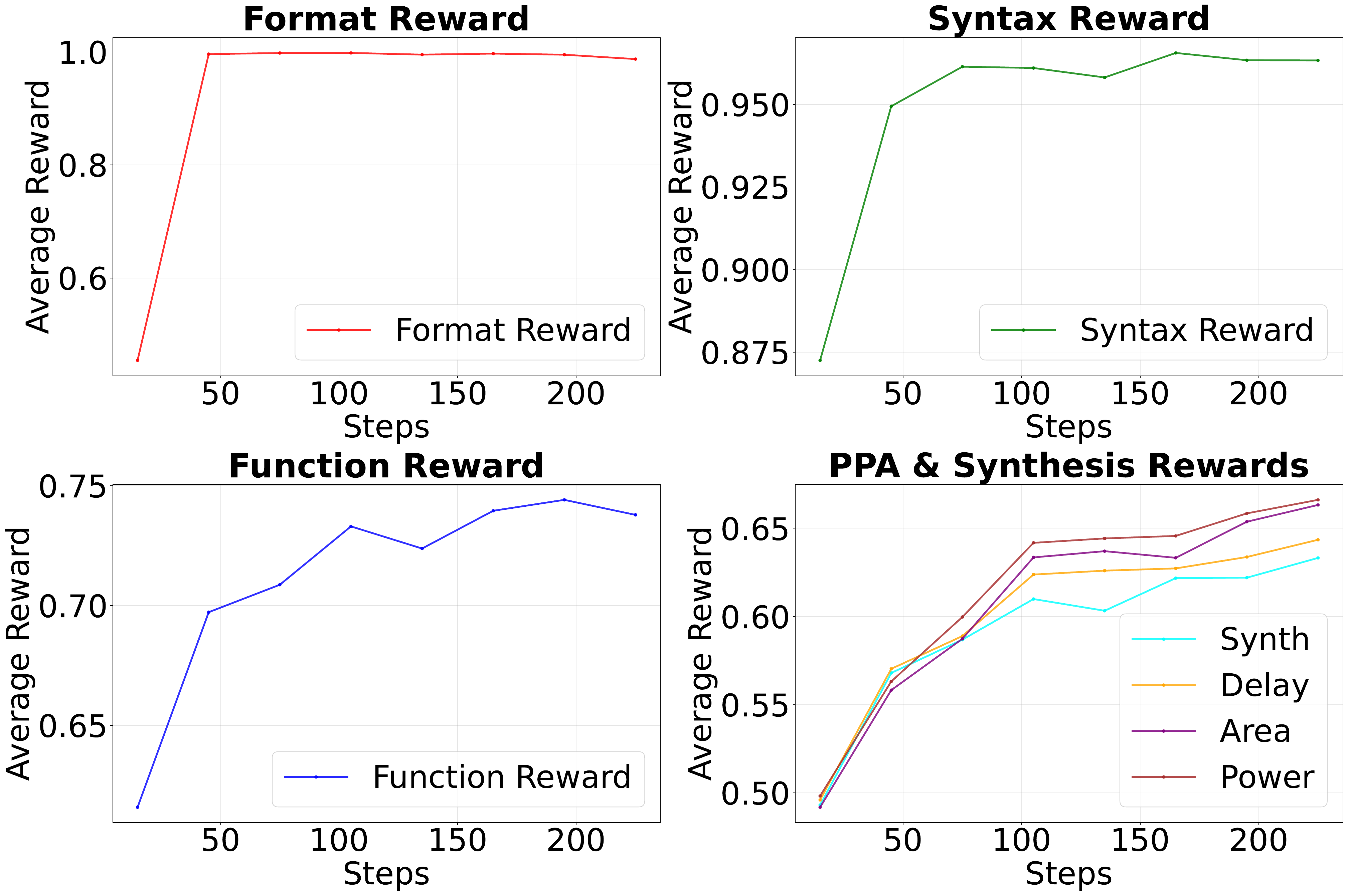}
    \caption{The Verilog code rewards and format reward increase during reinforcement learning. Each plotted point is computed as an average over a window of 30 neighboring steps to smooth the curves.
}
    \label{fig:rewards}
\end{figure}

\begin{table*}[t]

\centering
\setlength{\tabcolsep}{2.6pt} 
\renewcommand{\arraystretch}{0.95}
\small
\newcolumntype{Y}{>{\centering\arraybackslash}X}
\begin{tabularx}{\textwidth}{@{}>{\raggedright}p{2cm} l r *{6}{Y} *{2}{Y}@{}}
\toprule
\textbf{Type} & \textbf{Model} & \textbf{Size} & 
\multicolumn{3}{c}{\textbf{VerilogEval Machine (\%)}} & 
\multicolumn{3}{c}{\textbf{VerilogEval Human (\%)}} & 
\multicolumn{2}{c}{\textbf{RTLLM v1.1 (\%)}} \\
\cmidrule(lr){4-6} \cmidrule(lr){7-9} \cmidrule(lr){10-11}
 & & & 
\textbf{pass@1} & \textbf{pass@5} & \textbf{pass@10} & 
\textbf{pass@1} & \textbf{pass@5} & \textbf{pass@10} & 
\textbf{Syntax@5} & \textbf{Pass@5} \\
\midrule

\multirow{2}{*}{\parbox{2cm}{Foundational\\Models}} 
 & GPT-4o & - & 65.9 & 71.4 & 72.7 & 57.1 & 63.9 & 66.7 & 93.9 & 65.5 \\
 & GPT-o1 & - & 67.4 & 79.2 & 81.8 & 58.5 & 68.0 & 71.2 & 93.1 & 72.4 \\
\midrule

\multirow{4}{*}{Base Models} 
 & CodeLlama & 7B & 43.1  & 47.1 & 47.7 & 18.2 & 22.7 & 24.3 & 62.6 & 29.9 \\
 & DeepSeek-Coder & 6.7B & 52.2 & 55.4 & 56.8 & 30.2 & 33.9 & 34.9 & 64.4 & 29.3 \\
 & CodeQwen & 7B & 46.5 & 54.9 & 56.4 & 22.5 & 26.1 & 28.0 & 65.8 & 34.0 \\
 & Qwen2.5-Coder & 7B & 51.3 & 76.3 & 81.8 & 27.8 & 43.6 & 48.7 & 86.2 & 48.3 \\
\midrule

\multirow{1}{*}{Origen} 
 & DeepSeek-Coder & 7B & 74.1 & 82.4 & 85.7 & 54.4 & 60.1 & 64.2 & - & 65.5 \\
\midrule

\multirow{1}{*}{ReasoningV} 
 & Qwen2.5-Coder & 7B &  73.6 &  83.4 &  85.3 & 57.8 & 69.3 &  72.4 & - & 62.2 \\
\midrule

\multirow{2}{*}{RTLCoder } 
 & Mistral & 7B & 62.5 & 72.2 & 76.6 & 36.7 & 45.5 & 49.2 & 73.7 & 37.3 \\
 & DeepSeek-Coder & 7B & 61.2 & 76.5 & 81.8 & 41.6 & 50.1 & 53.4 & 83.9 & 40.3 \\
\midrule


\multirow{3}{*}{CodeV} 
 & CodeLlama & 7B & 78.1 & 86.0 & 88.5 & 45.2 & 59.5 & 63.8 & 89.2 & 50.3 \\
 & DeepSeek-Coder & 6.7B & 77.9 & 88.6 & 90.7 & 52.7 & 62.5 & 67.3 & 87.4 & 51.5 \\
 & CodeQwen & 7B & 77.6 & 88.2 & 90.7 & 53.2 & 65.1 & 68.5 & 89.5 & 53.3 \\
\midrule

\multirow{3}{*}{CraftRTL} 
 & CodeLlama & 7B & 78.1 & 85.5 & 87.8 & 63.1 & 67.8 & 69.7 & \underline{93.9} & 52.9 \\
 & DeepSeek-Coder & 6.7B & 77.8 & 85.5 & 88.1 & \underline{65.4} & 70.0 & 72.1 & 92.9 & 58.8 \\
 & Starcoder2 & 15B & 81.9 & 86.9 & 88.1 & \textbf{68.0} & \underline{72.4} & \underline{74.6} & \underline{93.9} & 65.8 \\

 \midrule

\multirow{4}{*}{\textbf{ChipSeek}} 
 & CodeLlama & 7B & \underline{85.7} & 88.8 & 89.5 & 63.4 & 70.1 & 72.4 & 93.1 & \textbf{82.8} \\
 & Deepseek-Coder & 7B & 83.3 & 88.9 & 90.2 & 64.3 & 71.1 & 73.7 & 93.1 & 72.4 \\
 & CodeQwen & 7B & \textbf{87.2} & \underline{90.3} & \underline{90.9} & 63.8 & 69.4 & 70.5 & 89.7 & \underline{75.8} \\
 & Qwen2.5-Coder & 7B & 84.1 & \textbf{90.6} & \textbf{92.3} & 62.2 & \textbf{73.7} & \textbf{76.9} & \textbf{96.6} & \textbf{87.2} \\
\bottomrule
\end{tabularx}
\caption{Evaluation Results on VerilogEval \cite{10.1145/3676536.3676830} and RTLLM v1.1 \cite{liu2024openllm}. We compare our models with GPT series, 4 coding language models, and several Verilog specific models including RTLCoder \cite{liu2024rtlcoderb}, CodeV \cite{zhao2024codevempoweringllmsverilog}, Origen \cite{10.1145/3676536.3676830}, CraftRTL  \cite{liu2025craftrtlhighqualitysyntheticdata} and ReasoningV \cite{qin2025reasoningvefficientverilogcode}.}
\label{function_compare}
\end{table*}

\newcolumntype{Y}{>{\centering\arraybackslash}X}


\begin{table}[t]

    \centering
    \small
    \setlength{\tabcolsep}{8pt} 
    \renewcommand{\arraystretch}{1}

    \newcolumntype{Y}{>{\centering\arraybackslash}X}

    \begin{tabular*}{\linewidth}{@{\extracolsep{\fill}}lccccc}
        \toprule
        \multirow{2}{*}{\textbf{Model}}  & 
        \multicolumn{1}{c}{\textbf{Func.} $\uparrow$} & 
        \multicolumn{3}{c}{\textbf{EDAP} $\downarrow$}   \\

        \cmidrule(lr){2-2} \cmidrule(lr){3-5} 
        
        & \textbf{pass@5}  & \textbf{max} & \textbf{avg} & \textbf{min}  \\
        \midrule
        GPT-4o & 70.45 & 1.13 & 1.02 & 0.84  \\ 
        GPT-o1  & 72.73 & 1.14 & 1.06 & 0.99 \\ 
        \midrule
        CodeLlama & 43.18 & 1.04 & 0.94 & 0.87 \\ 
        DeepSeek-Coder-7B & 56.82 & 3.00 & 2.65 & 1.91 \\ 
        CodeQwen & 38.64 & 2.76 & 2.56 & 2.41 \\ 
        Qwen2.5-Coder & 56.82 & 1.38 & 1.04 & 0.89 \\ 
        \midrule
        Verigen-MCTS & 45.45 & 0.96 & 0.93 & 0.87 \\
        \midrule
        RTLCoder-M  & 50.00 & 2.97 & 2.43 & 0.83 \\ 
        RTLCoder-DC-6.7B & 52.27 & 5.83 & 3.73 & 1.41 \\ 
        \midrule
        CodeV-CL & 59.09 &  2.12 & 1.43 & 1.02 \\ 
        CodeV-DC-6.7B & 63.64 & 2.30 & 1.56 & 1.28 \\ 
        CodeV-CQ  & 61.36 & 3.35 & 2.53 & 1.03  \\ 
        \midrule
        Origen-DC-7B   & 65.91 & 3.17 & 1.45  & 0.85 \\ 
        \midrule
        ChipSeek-CL  & \underline{77.27} & \underline{0.85} & \textbf{0.76} & \textbf{0.74} \\ 
        ChipSeek-DC-7B & 75.00 & 0.93 & 0.84 & \underline{0.78} \\ 
        ChipSeek-CQ & 72.73 & \textbf{0.84} & \underline{0.83} & 0.81 \\ 
        ChipSeek-QC  & \textbf{84.09} & 0.96 & 0.88 & 0.81 \\ 

        \bottomrule
    \end{tabular*}
    \caption{A comparison of various models and methods on function and overal performance. EDAP values are normalized to the original benchmark. QC: Qwen2.5 Coder, M: Mistral, DC: DeepSeek-Coder, CQ: CodeQwen, CL: CodeLlama}
    \label{ppa_compare}
\end{table}

\begin{table}[h!]

    \centering
    \small
    \setlength{\tabcolsep}{1.6pt} 
    \renewcommand{\arraystretch}{1.3} 

    \begin{tabular}{lcccccc}
        \hline
        \textbf{Model} & \textbf{Delay$\downarrow$} & \textbf{Area$\downarrow$} & \textbf{Power$\downarrow$} & \textbf{ADP$\downarrow$} & \textbf{EDP$\downarrow$} & \textbf{Error$\downarrow$} \\
        \hline
        
        Base Model & 0.93 & 1.02 & 1.02 & 1.02 & 1.02 & 0.46  \\
        Origen & 0.88 & 0.93  & 0.93  & 0.85 & 0.86 &0.34        \\
        ChipSeek & 0.83 & 0.88 & 0.85 & 0.80 & 0.77 & 0.25 \\
        \hline
    \end{tabular}
        \caption{Specific design goals optimization capability of Verilog Coding Models with the DeepSeek-Coder-v1.5-7B as the Base Model.}
        \label{tab:dseek_results}
\end{table}

\section{Experiment}
\label{experiment}
\subsection{Implementation Details}
\label{Implement}

We adopt the CodeLlama-7B, CodeQwen-7B, DeepSeek-Coder-v1.5-7B and Qwen2.5-Coder-7B-Instruct model as our base models. During the cold-start fine-tuning stage, we use \textit{29,127} data samples to impart preliminary reasoning and Verilog generation capabilities to the model. Subsequently, during reinforcement learning stage, we further train the model using \textit{8,453} data samples to enhance its capabilities. Both training datasets are strictly filtered to exclude any cases from the benchmark. We conduct all training processes on a cluster of 4 NVIDIA A100 80GB GPUs, leveraging the DeepSpeed distributed training framework and the vLLM inference framework to accelerate training.


Furthermore, we employ a multi-faceted strategy to accelerate the computationally intensive reward calculation process. First, we utilize a thread pool to concurrently run multiple simulation and synthesis tasks. Additionally, we implement a hierarchical gating technique based on a ``fail-fast" principle. In this pipeline, any Verilog sample that fails an early-stage evaluation is immediately discarded and not passed to subsequent more costly stages. For example, a design that fails the functional correctness simulation will not proceed to the PPA evaluation. This multistage filtering dramatically reduces computation time by avoiding full evaluations for infeasible candidates, significantly accelerating the overall reward calculation process.

As shown in Figure \ref{fig:rewards}, the rewards designed in Section \ref{sec:rewards} steadily increase with training steps. The rise in format reward indicates that the model has learned to apply chain of thought before code generation. The increase in Verilog code reward reflects the model's improved chip design capability during reinforcement learning, including enhanced syntax correctness, functional correctness, and PPA performance. This upward trend in rewards provides preliminary evidence for the effectiveness of our method.








\subsection{Results}

\paragraph{Functional Correctness:} In Table \ref{function_compare}, we compare the Verilog functional correctness of our proposed model against several baseline models on RTLLM v1.1 and VerilogEval benchmarks. On the VerilogEval benchmark, our model ChipSeek achieves the best performance for \textit{pass@1}, \textit{pass@5}, and \textit{pass@10} in the Machine track. In the Human track, it attains state-of-the-art results for \textit{pass@5} and \textit{pass@10}, and its performance for \textit{pass@1} is comparable to the previous best. On the RTLLM v1.1 benchmark, our generated code achieves the highest \textit{pass@5} rate in both functionality and syntactical correctness, surpassing the previous best by 21.4\% and 2.7\% respectively. In Table \ref{function_compare}, we highlight the best scores in boldface.

\paragraph{Overall Performance Evaluation:} To compare the core design capabilities of different models, we first conduct an overall performance evaluation on RTLLM v2.0 without any specified optimization objectives, with the results shown in Table \ref{ppa_compare}. We focus our comparison on two key metrics: pass@5 and the comprehensive Energy-Delay-Area Product (EDAP). We evaluate the original benchmark, GPT series, RTLCoder series, CodeV series, Origen, Verigen-MCTS and our ChipSeek series. For each design prompt, every large language model generates 10 candidate solutions. We then select the functionally correct designs and report the maximum, mean, and minimum EDAP scores among these candidates. Our ChipSeek models achieve state-of-the-art results, with our best configuration improving functional correctness (pass@5) by 11.4\% and reducing the worst, average and best EDAP by 20\%, 18\% and 9\% respectively compared to the previous best-performing models. 

\begin{figure}[!t] 
    \centering
    \includegraphics[width=\linewidth]{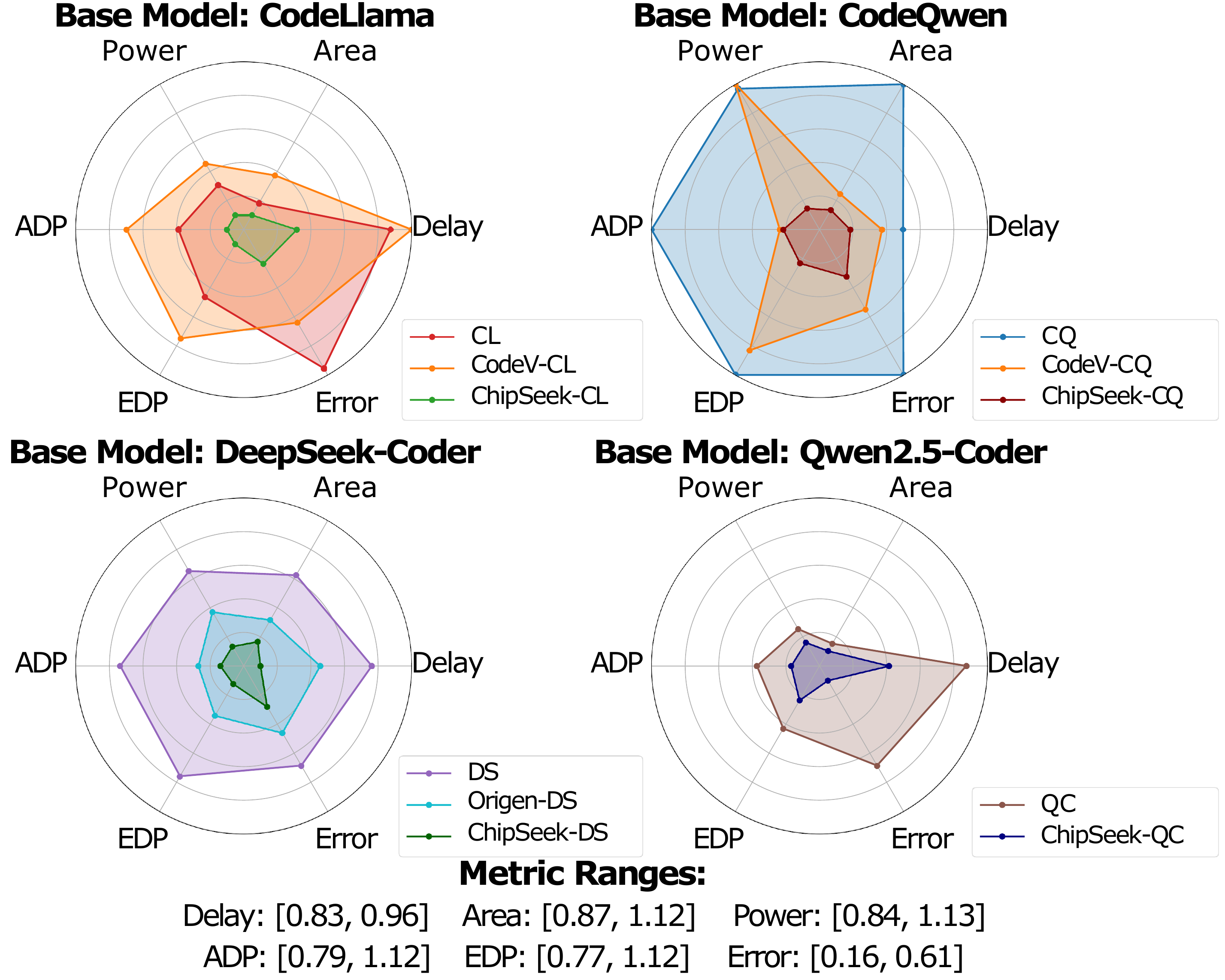}
    \caption{Radar chart comparison of key performance metrics. 
}
    \label{fig:radar_chart}
\end{figure}




\paragraph{Performance Evaluation With Design Priorities:} To investigate the model's fine-grained design capabilities, we augmented the RTLLM v2.0 benchmark with specific optimization objectives communicated through targeted prompts. We measure the metrics including power, delay, area, ADP (Area-Delay Product), EDP (Energy-Delay Product) and functional error rate across 11 models grouped by 4 base models, with the comprehensive results visualized in the radar charts in Figure \ref{fig:radar_chart}. We choose the best metric score across 5 attempts for each problem.  Each axis on the charts represents a normalized score, where values closer to the center signify better performance. The visual evidence shows that the polygon representing the ChipSeek model is substantially smaller and more centered than that of its corresponding baseline in every comparison, demonstrating superior results across all metrics simultaneously.

For a quantitative view, Table \ref{tab:dseek_results} details the results on the DeepSeek-Coder-7B based models. Compared to the strong Origen baseline, our ChipSeek method significantly improves all metrics compared to the base model, reducing delay by 5\%, area by 5\%, power by 8\%, ADP by 5\% and EDP by 9\%. This comprehensive improvement highlights the sophisticated trade-off navigation learned through our dynamic preference reinforcement learning framework.

\begin{figure}[!t] 
    \centering
    \includegraphics[width=\linewidth]{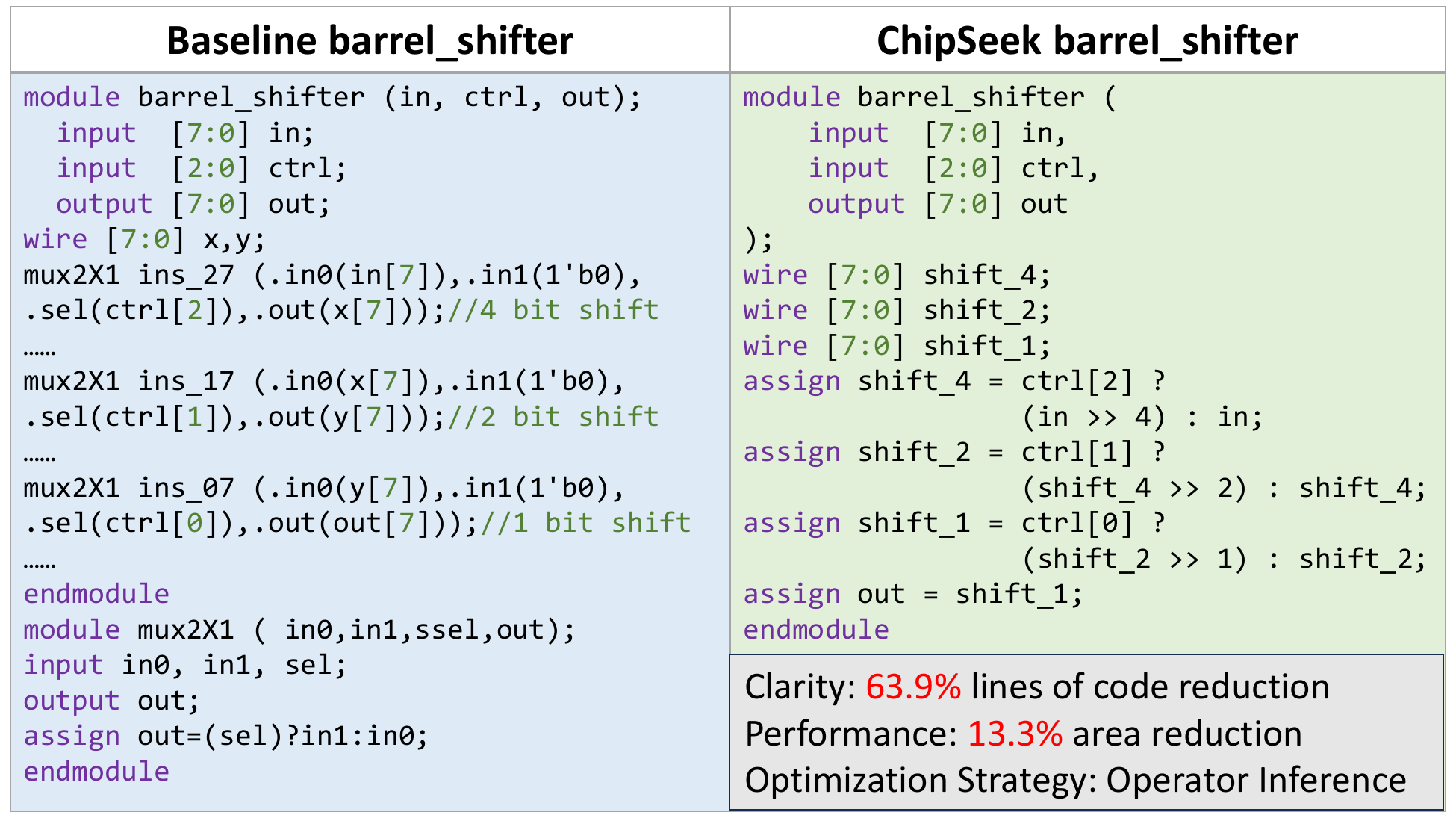}
    \caption{Case Study of ChipSeek. 
}
    \label{fig:case_study}
\end{figure}

\paragraph{Case Study:} The barrel shifter is a crucial component in high-performance computing. As shown on the left in Figure \ref{fig:case_study}, the baseline design uses a traditional implementation with explicitly instantiated multiplexer (MUX) sub-modules. In contrast, shown on the right, our model found that describing only the high-level barrel-shift behavior—without employing MUX sub-modules—allows backend EDA tools to perform the optimization of operator inference. This process, combining a large-scale generation of candidate circuits with rapid RL-based iteration and EDA feedback, yields a powerful co-optimization of front-end and back-end design stages.

\begin{figure}[!t] 
    \centering
    \includegraphics[width=\linewidth]{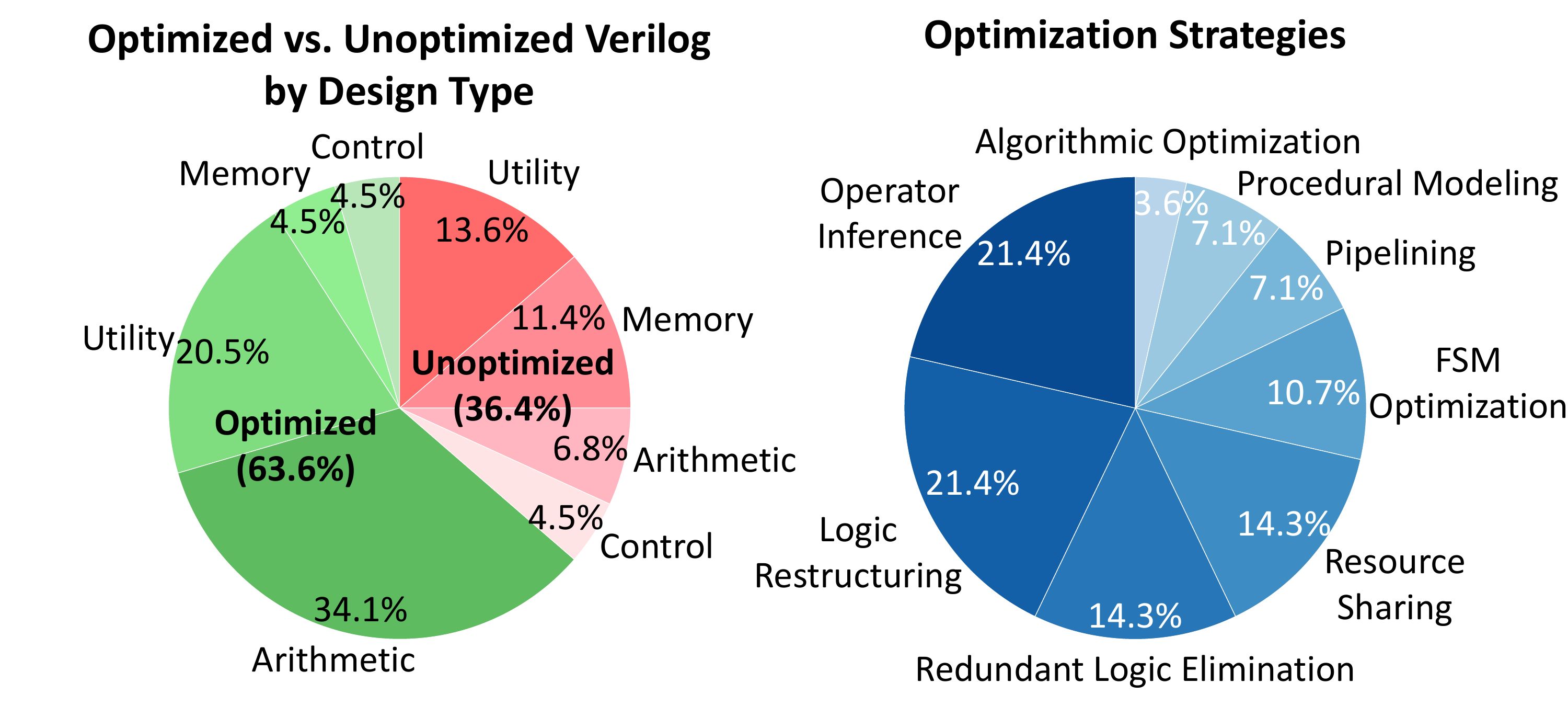}
    \caption{Optimization Analysis of ChipSeek. 
}
    \label{fig:distribution}
\end{figure}

\begin{table}[t!]

    \centering
    \small
    \setlength{\tabcolsep}{3.8pt} 
    \renewcommand{\arraystretch}{1.3} 

    \begin{tabular}{lccccccccc}
        \hline
        \textbf{Method} & \textbf{Syntax$\uparrow$} & \textbf{Pass@1$\uparrow$} & \textbf{Synthesis$\uparrow$} & \textbf{EDAP$\downarrow$} \\
        \hline
        
        ChipSeek & 91.36\% & 71.36\% & 70.91\% & 0.81 \\
        \hline 
        w/o Format & -7.27\% & -10.00\% & -10.23\% & +0.08 \\
        w/o Syntax & -2.73\% & -0.45\%  & -0.68\%  & +0.02         \\
        w/o Function & -0.91\% & -14.31\% & -14.55\% & +0.56 \\
        w/o Synthesis & +0.45\% & -0.22\% & -3.18\% & +0.05 \\
        w/o PPA & -0.68\% & -1.36\%  & -1.36\% & +1.10 \\
        
        \hline
    \end{tabular}
    \caption{Ablation Study of Rewards. We use Qwen2.5-Coder-7B as the base model to conduct the experiment.}
    \label{tab:ChipSeek_ablation_compact_final}
\end{table}

\begin{table}[t!]

    \centering
    \small
    \setlength{\tabcolsep}{5pt} 
    \renewcommand{\arraystretch}{1.1} 

    \begin{tabular}{lccccccccc}
        \hline
        \textbf{Method}  & \textbf{Delay$\downarrow$} & \textbf{Area$\downarrow$} & \textbf{Power$\downarrow$} & \textbf{ADP$\downarrow$} & \textbf{EDP$\downarrow$} \\
        \hline
        
        GRPO & 0.89 & 0.90 & 0.91 & 0.85 & 0.85 \\
        DAPO & 0.92 & 0.90 & 0.89 & 0.87 & 0.86 \\
        CDPO & 0.84 & 0.87 & 0.86 & 0.79  & 0.77\\
        
        \hline
    \end{tabular}
    \caption{Ablation Study of Reinforcement Learning Methods \cite{shao2024deepseekmathpushinglimitsmathematical, yu2025dapoopensourcellmreinforcement} on different RTL Performance Optimization goals.}
    \label{tab:policy_ablation_study}
\end{table}

\subsection{Ablation Study}

We conducted an ablation study on the RTLLM v2.0 to evaluate the contribution of each component within our framework, as detailed in Table \ref{tab:ChipSeek_ablation_compact_final}. In this study, we systematically removed each reward and measured the impact on a suite of metrics: syntax pass rate, pass@5, synthesis pass rate, and EDAP. The results in Table \ref{tab:ChipSeek_ablation_compact_final} show that the removal of any single component leads to performance degradation to varying degrees. Furthermore, we compared our CDPO algorithm with prior optimization methods. As shown in Table \ref{tab:policy_ablation_study}, CDPO consistently outperforms both GRPO and DAPO across all RTL performance goals, highlighting its superior capabilities in multi-objective optimization tasks for critical chip design metrics.
    
\begin{figure}[!t] 
    \centering
    \includegraphics[width=\linewidth]{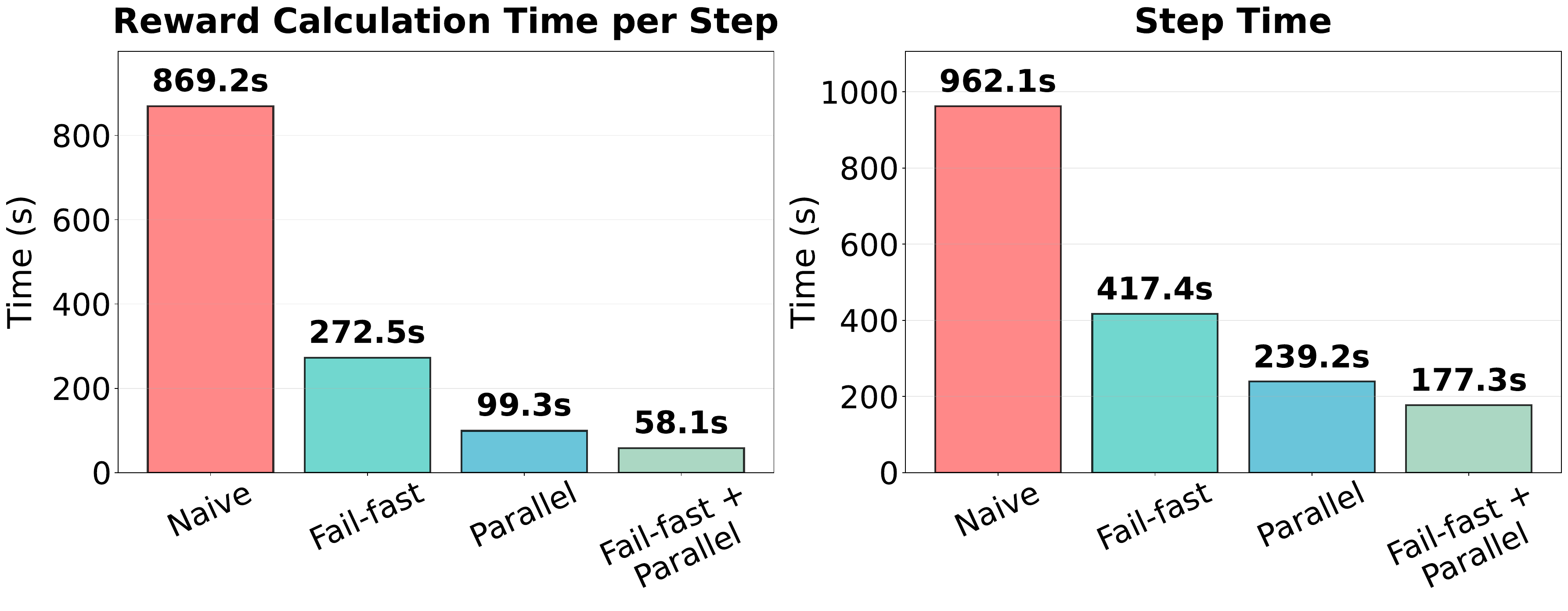}
    \caption{Effect of Acceleration on Training Time.
}
    \label{fig:runtime_analysis}
\end{figure}

\subsection{Runtime Analysis}
Compact and fast-inference RTL generation models are essential for efficient chip design work flows, particularly given the computational intensity and resource constraints of large-scale designs. Utilizing the acceleration techniques detailed in Section 4.1, we significantly reduce computational overhead, achieving marked improvements in runtime
performance. Specifically, as demonstrated in Figure \ref{fig:runtime_analysis}, we achieve a substantial reduction of 93.5\% in reward calculation time and 81.6\% in step update time. To ensure a rigorous and fair evaluation, all experiments were executed with
the same training config on an isolated, dedicated server without concurrent processes.

\section{Optimization Analysis}
Shown in Figure \ref{fig:distribution}, the optimization profile of ChipSeek demonstrates a strong competency in gate-level and structural refinement. The model predominantly leverages techniques such as Operator Inference and Logic Restructuring (both at 21.4\%) to enhance design efficiency. This proficiency is particularly impactful on Arithmetic modules, which constitute the largest portion of optimized designs. However, despite these strengths, the model exhibits a comparatively weaker optimization capability on Control and Memory modules. We hypothesize that this stems from a relative scarcity of these circuit types within the training dataset, which may have limited the model's exposure to their unique optimization patterns.

\section{Conclusion}

We introduce ChipSeek, a reinforcement learning framework for superior RTL generation. By using direct feedback from the EDA toolchain as reward signals within our CDPO algorithm, ChipSeek simultaneously optimizes for both functional correctness and specific PPA goals. On standard benchmarks, this approach allows our model to significantly outperform previous models, demonstrating superior results in both functional correctness and PPA performance. ChipSeek thus enables more efficient automated hardware design by bridging the gap between the training process and EDA toolchain feedback.

\section*{Limitations}
Our approach still has several practical limitations. First, although the closed-loop design enables direct optimization against compilation, simulation, and synthesis feedback, it remains computationally heavy due to repeated EDA invocations; in particular, reward evaluation still takes a long time even with the proposed gating mechanism to skip unnecessary tool calls. Second, while the optimization profile indicates strong competency in gate-level and structural refinement, the gains are not uniform across circuit categories: consistent with our optimization analysis, ChipSeek exhibits comparatively weaker optimization capability on \textit{Control} and \textit{Memory} modules than on \textit{Arithmetic} and \textit{Utility} modules. We hypothesize that this gap is partly driven by a relative scarcity of control- and memory-centric designs in the training/evaluation data, which limits the model's exposure to their distinct optimization patterns (e.g., stateful control flow and memory access behaviors) and makes it harder to consistently discover effective refinements under the same feedback budget.

\bibliography{custom}

\clearpage

\appendix

\section{CDPO Algorithm}
\label{sec:appendix}
\begin{algorithm}[h]
\caption{CDPO (Curriculum-Guided Dynamic Policy Optimization)}
\label{alg:cdpo_short}
\begin{algorithmic}[1]
\REQUIRE Policy $\pi_{\theta}$, behavior policy $\pi_{\theta_{\text{old}}}$; batch size $B$, group size $G$;
process rewards $\mathcal{R}_{pro}$, objective rewards $\mathcal{R}_{cor}=\{\text{func},P,D,A\}$;
EMA factor $\beta$, $\epsilon$, clip bounds $\varepsilon_{\text{low}},\varepsilon_{\text{high}}$, $w_{func}$;
curriculum weights $\{\alpha_k\}_{k\in\mathcal{R}_{pro}}$
\ENSURE Updated $\theta$ and $\{\alpha_k\}$

\FOR{training step $s=1,2,\dots$}
    \STATE Sample prompts $\{q_j\}_{j=1}^{B}$
    \FOR{each prompt $q_j$}
        \STATE $\mathbf{p}(q_j)\leftarrow \textsc{ExtractPref}(q_j)$
        \STATE Sample $\{o_{j,i}\}_{i=1}^{G}\sim \pi_{\theta_{\text{old}}}(\cdot\mid q_j)$
        \STATE $\{r^{(k)}_{j,i}\}\leftarrow \textsc{EvalToolchain}(q_j,o_{j,i})$
        \STATE $\{\widehat{A}^{(k)}_{j,i}\}\leftarrow \textsc{Advantage}(\{r^{(k)}_{j,i}\}_{i=1}^{G},\epsilon)$
    \ENDFOR

    \FOR{each $k\in\mathcal{R}_{pro}$}
        \STATE $\bar{\mu}^{(s)}_k \leftarrow \frac{1}{BG}\sum_{j=1}^{B}\sum_{i=1}^{G} r^{(k)}_{j,i}$
        \STATE $\hat{\alpha}^{(s)}_k \leftarrow \max(0,1-\bar{\mu}^{(s)}_k)$
        \STATE $\alpha^{(s)}_k \leftarrow \beta\,\alpha^{(s-1)}_k + (1-\beta)\,\hat{\alpha}^{(s)}_k$
    \ENDFOR

    \STATE $\mathcal{L}\leftarrow 0$
    \FOR{each token $o_{j,i,t}$ in all rollouts}
        \STATE $A^{pro}\leftarrow \sum_{k\in\mathcal{R}_{pro}}\alpha^{(s)}_k\,\widehat{A}^{(k)}_{j,i}$
        \STATE $A^{cor}\leftarrow w_{func}\widehat{A}^{(func)}_{j,i} + \sum_{m\in\{P,D,A\}} p_m(q_j)\widehat{A}^{(m)}_{j,i}$
        \STATE $A^{total}\leftarrow A^{pro}+A^{cor}$
        \STATE $r(\theta)\leftarrow \frac{\pi_{\theta}(o_{j,i,t}\mid q_j,o_{j,i,<t})}{\pi_{\theta_{\text{old}}}(o_{j,i,t}\mid q_j,o_{j,i,<t})}$
        \STATE $\tilde r \leftarrow \textsc{Clip}(r,1-\varepsilon_{\text{low}},1+\varepsilon_{\text{high}})$
        \STATE $\mathcal{L}\leftarrow \mathcal{L} - \min\!\big(rA^{total},\tilde rA^{total}\big)$
    \ENDFOR
    \STATE Update $\theta$ by minimizing $\mathcal{L}$; set $\theta_{\text{old}}\leftarrow \theta$ \hfill \textit{(RL step)}
\ENDFOR
\end{algorithmic}
\end{algorithm}

\begin{table*}[h]
    \centering
    \renewcommand{\arraystretch}{1}
    \newcolumntype{Y}{>{\centering\arraybackslash}X}
    \begin{tabular*}{\linewidth}{@{\extracolsep{\fill}}lcccccc}
        \toprule
        \multirow{2}{*}{\textbf{Model}}  & 
        \multicolumn{3}{c}{\textbf{VerilogEval-Machine(\%)}} & 
        \multicolumn{3}{c}{\textbf{VerilogEval-Human(\%)}}   \\
        \cmidrule(lr){2-4} \cmidrule(lr){5-7} 
        
        & \textbf{pass@1} & \textbf{pass@5} & \textbf{pass@10}  & \textbf{pass@1} & \textbf{pass@5} & \textbf{pass@10}  \\
        \midrule
        CodeLlama &  43.1  & 47.1 & 47.7 & 18.2 & 22.7 & 24.3  \\
        +SFT  & 48.1 & 60.2 & 69.9 & 27.8 & 53.6 & 58.9  \\
        +SFT+CDPO  & 85.7 & 88.8 & 89.5 & 63.4 & 70.1 & 72.4  \\
        \midrule
        DeepSeek-Coder &  52.2 & 55.4 & 56.8 & 30.2 & 33.9 & 34.9  \\
        +SFT  & 55.7 & 75.4 & 78.1 & 33.8 & 45.8 & 52.3 \\
        +SFT+CDPO & 83.3 & 88.9 & 90.2 & 64.3 & 71.1 & 73.7  \\
        \midrule
        CodeQwen  & 46.5 & 54.9 & 56.4 & 22.5 & 26.1 & 28.0  \\
        +SFT & 51.5 & 78.9 & 82.3 & 31.9 & 49.5 & 53.8 \\
        +SFT+CDPO & 87.2 & 90.3 & 90.9 & 63.8 & 69.4 & 70.5 \\
        \midrule
        Qwen2.5-Coder & 51.3 & 76.3 & 81.8 & 27.8 & 43.6 & 48.7  \\
        +SFT  & 57.3 & 75.4 & 79.0 & 34.7 & 49.9 & 54.5 \\
        +SFT+CDPO  & 84.1 & 90.6 & 92.3 & 62.2 & 73.7 & 76.9 \\
        \bottomrule
    \end{tabular*}
    \caption{An ablation study of Training Stage on VerilogEval. }
    \label{ablation_study_verilogeval}
\end{table*}

\begin{table*}[h]
    \centering
    \renewcommand{\arraystretch}{1}
    \newcolumntype{Y}{>{\centering\arraybackslash}X}
    \begin{tabular*}{\linewidth}{@{\extracolsep{\fill}}lcccccccc}
        \toprule
        \multirow{2}{*}{\textbf{Model}}  & 
        \multicolumn{2}{c}{\textbf{RTLLM v2.0-Func (\%)}} & 
        \multicolumn{6}{c}{\textbf{RTLLM v2.0-Performance}}   \\
        \cmidrule(lr){2-3} \cmidrule(lr){4-9} 
        
        & \textbf{Syntax} & \textbf{pass@5}  & \textbf{Power} & \textbf{Area} & \textbf{Delay} & \textbf{ADP} & \textbf{EDP} & \textbf{EDAP}  \\
        \midrule
        CodeLlama &  36.4  & 34.1 & 0.93 & 0.92 & 0.99 & 0.95& 0.96 & 0.94  \\
        +SFT  & 78.2 & 62.3 & 1.20 & 1.02 & 1.04 & 1.09 & 1.25 & 1.47  \\
        +SFT+CDPO  & 89.5 & 77.3 & 0.87 & 0.91 & 0.91 & 0.83 & 0.80 & 0.76  \\
        \midrule
        DeepSeek-Coder &  67.5 & 50.1 & 1.13 & 1.13 & 1.00 & 1.17 & 1.17 & 2.65  \\
        +SFT & 80.2 & 64.0 & 1.05 & 1.04 & 1.00 & 1.06 & 1.08 & 1.51  \\
        +SFT+CDPO & 88.6 & 75.0 & 0.91 & 0.93 & 0.83 & 0.90 & 0.88 & 0.84  \\
        \midrule
        CodeQwen  & 33.8 & 33.8 & 1.18 & 1.18 & 0.95 & 1.20 & 1.21 & 2.56  \\
        +SFT & 81.6 & 69.1 & 1.05 & 1.06 & 1.02 & 1.10 & 1.10 & 1.42 \\
        +SFT+CDPO  & 87.3 & 72.7 & 0.89 & 0.92 & 0.94 & 0.88 & 0.86 & 0.82 \\
        \midrule
        Qwen2.5-Coder & 67.0 & 53.1 & 0.96 & 0.95 & 1.03 & 1.01 & 1.01 & 1.04  \\
        +SFT  & 79.5 & 68.5 & 1.07 & 1.03 & 1.05 & 1.10 & 1.15 & 1.31 \\
        +SFT+CDPO  & 91.4 & 84.1 & 0.93 & 0.92 & 0.95 & 0.91 & 0.92 & 0.88 \\
        \bottomrule
    \end{tabular*}
    \caption{An ablation study of Training Stage on RTLLM-v2.0. Energy, Delay, Area, ADP, EDP, EDAP are all normalized to the benchmark baseline.}
    \label{ablation_rtllm}
\end{table*}

\begin{table*}[t]
    \centering
    \renewcommand{\arraystretch}{1}
    \newcolumntype{Y}{>{\centering\arraybackslash}X}
    \begin{tabular*}{\linewidth}{@{\extracolsep{\fill}}lccccccccc}
\toprule
Method & Syntax & Pass@5 & Synthesis &  Power & Area & Delay & ADP & EDP & EDAP \\
\midrule
ChipSeek   & 91.4 & 84.1 & 70.91 & 0.93 & 0.92 & 0.95 & 0.91 & 0.92 & 0.88 \\
-AdvAgg       & 88.7 & 78.1 & 68.8 & 0.95 & 0.94 & 0.96 & 0.93 & 0.94 & 0.92 \\
-Curriculum   & 90.8 & 72.5 & 63.2 & 0.96 & 0.95 & 0.95 & 0.94 & 0.94 & 0.93 \\
-PreVec       & 91.2 & 84.0 & 70.1 & 0.95 & 0.93 & 0.96 & 0.94 & 0.95 & 0.93 \\
\bottomrule
    \end{tabular*}
\caption{Ablation study on the effectiveness of key components in ChipSeek.
\textbf{AdvAgg} denotes \emph{Advantage-Level Aggregation}, \textbf{Curriculum} denotes the \emph{Curriculum Weight Schedule}, and \textbf{PreVec} denotes the \emph{Preference Vector}.} 
\label{tab:ablation_main}
\end{table*}

\section{Implementation Details}

ChipSeek implements a comprehensive multi-stage reward evaluation system specifically designed for Verilog code generation with PPA (Power, Performance, Area) optimization using Dynamic Preference Policy Optimization (CDPO). The reinforcement learning algorithm is built upon VeRL \cite{verl}. The core architecture consists of a specialized reward manager that orchestrates the evaluation of generated Verilog designs across multiple dimensions including functional correctness, synthesis feasibility, and PPA performance metrics. The system incorporates several critical training speed optimizations including DeepSpeed ZeRO-2 acceleration strategy for efficient distributed training and vLLM for high-throughput inference generation.

The functional correctness evaluation phase utilizes Icarus Verilog simulation within containerized environments to assess both syntactic validity and behavioral correctness of generated designs. Each Verilog code snippet is compiled with comprehensive warning flags and executed against provided testbenches to determine functional accuracy. This phase produces binary rewards for compilation success and testbench passage, establishing a foundational requirement that designs must be functionally correct before proceeding to performance evaluation. To optimize training efficiency, the system implements intelligent filtering mechanisms that remove training samples where all generated responses have identical reward values, thereby focusing computational resources on diverse and informative examples.

For designs that pass functional verification, the system performs comprehensive PPA analysis using Yosys \cite{Wolf2013YosysAFV} and OpenROAD \cite{openroad} in a secondary evaluation phase.  The synthesis process extracts detailed power consumption, area utilization, and timing performance metrics from the generated designs. A key innovation lies in the dynamic preference-based evaluation mechanism, where user-specified preference vectors allow flexible weighting of different PPA objectives, enabling the system to optimize for power-efficient, area-efficient, or performance-optimized designs based on application requirements. The reward manager implements efficient parallel processing with ThreadPoolExecutor to handle batch evaluation of multiple design candidates simultaneously, significantly reducing evaluation time during training. All third-party benchmarks/tools are used in accordance with their original licenses, and we will release our code and scripts under an open-source license.

The training configuration employs carefully tuned hyperparameters optimized for Verilog code generation tasks. The system uses asymmetric clipping ratios with clip\_ratio\_low=0.2 and clip\_ratio\_high=0.28 to provide more flexibility for positive policy updates while maintaining strict constraints on negative updates. Training operates with a batch size of 32 prompts, generating 8 responses per prompt, and uses a conservative learning rate of 1e-6 with 10 warmup steps and weight decay of 0.1. The generation process employs high diversity settings with temperature=1.0 and top\_p=1.0 during training, while validation uses more conservative top\_p=0.7 for stable evaluation. Sequence lengths are configured for max\_prompt\_length=2048 and max\_response\_length=8192 tokens to accommodate complex Verilog design descriptions.

Memory and computational efficiency optimizations include FSDP (Fully Sharded Data Parallel) with parameter and optimizer offloading enabled, sequence parallelism with sp\_size=2, and dynamic batch sizing to maximize GPU utilization. The system uses vLLM \cite{vllm} with tensor model parallelism (gen\_tp=2) for inference acceleration, chunked prefill for memory efficiency, and gradient checkpointing to reduce memory footprint during training. Additionally, an overlong buffer mechanism with penalty\_factor=1.0 discourages excessively long responses that could impact training stability. We use 13 x 4 A100 GPU hours to fully train our models.

Response format enforcement is integrated through structured output validation, requiring generated content to follow a reasoning-then-answer format with explicit thinking and solution sections. The final reward signal combines compilation feasibility, functional correctness, PPA improvement ratios relative to reference implementations, synthesis feasibility, and format compliance into a comprehensive score that guides the CDPO training process. This multi-faceted approach enables ChipSeek to learn nuanced trade-offs between different design objectives while maintaining computational efficiency through strategic filtering and acceleration techniques.

\section{Evaluation Details}

For functionality evaluation, we use the unbiased pass@k metrics to evaluate the functional correctness of the generated designs, calculated as Equation \ref{passk}. This metric estimates the probability that at least one functionally correct design is generated out of $k$ attempts. $n$ represents the total number of generations and $c$ represents the number of successfully generated code.

\begin{equation}
    \label{passk}
    \text{pass}@k := \mathbb{E}_{\text{task}} \left[ 1 - \frac{{\binom{n - c}{k}}}{{\binom{n}{k}}} \right],
\end{equation}

For performance evaluation of a design, we use fine-grained metrics including Delay, Area, Power and comprehensive metrics including ADP (Area-Delay-Product), EDP (Energy-Delay-Product) and EDAP (Energy-Delay-Area Product), calculated as Equation \ref{eq:adp}, \ref{eq:edp}, \ref{eq:edap}.

\begin{equation}
    \label{eq:adp}
    \text{ADP}_{\text{gen}} = \text{area}_{\text{gen}} \times \text{delay}_{\text{gen}}
\end{equation}

\begin{equation}
    \label{eq:edp}
    \text{EDP}_{\text{gen}} = \text{delay}_{\text{gen}} \times \text{power}_{\text{gen}}
\end{equation}

\begin{equation}
    \label{eq:edap}
    \text{EDAP}_{\text{gen}} = \text{area}_{\text{gen}} \times \text{delay}_{\text{gen}} \times \text{power}_{\text{gen}}
\end{equation}

For an overall performance evaluation of a model, we first filter the entire set of generated designs for each task, $\mathcal{G}_d$, to retain only those that are functionally correct and pass the testbench. This creates a filtered set of valid designs, shown in Equation \ref{eq:filter}.

\begin{equation}
    \label{eq:filter}
    S_d = \{ s \in \mathcal{G}_d \mid s \text{ passes the testbench} \}
\end{equation}

For each valid design $s \in S_d$, we then normalize its performance metric, $\text{Perf}(s)$, against the reference design's performance, $\text{Perf}_{\text{ref}}(d)$, for that task, calculating the normalized score as Equation \ref{eq:norm}.

\begin{equation}
    \label{eq:norm}
    \text{Perf}_{\text{norm}}(s) = \text{Perf}(s) / \text{Perf}_{\text{ref}}(d)
\end{equation}

where
\begin{equation}
    \label{eq:perf}
    \text{Perf} \in \{\text{Delay}, \text{Area}, \text{Power}, \text{ADP}, \text{EDP}, \text{EDAP}\}
\end{equation}

From this set of normalized scores for each task, we select the best (minimum), average, and worst (maximum) candidates to characterize the model's performance range on that specific task. The scores for each task $d$ are calculated as follows:
\begin{align}
    \label{eq:task_best}
    \text{Score}_{\text{best}}(d) &= \min_{s \in S_d} \{ \text{Perf}_{\text{norm}}(s) \} \\
    \label{eq:task_avg}
    \text{Score}_{\text{avg}}(d) &= \frac{1}{|S_d|} \sum_{s \in S_d} \text{Perf}_{\text{norm}}(s) \\
    \label{eq:task_worst}
    \text{Score}_{\text{worst}}(d) &= \max_{s \in S_d} \{ \text{Perf}_{\text{norm}}(s) \}
\end{align}
Finally, the overall evaluation score for the model is determined by averaging these individual task scores across all successfully generated tasks $\mathcal{T}$ in the benchmark. This provides a comprehensive view of the model's capabilities, captured in the final metric:
\begin{equation}
    \label{eq:overall_score}
    \text{Perf}_{\text{method}} = \frac{1}{|\mathcal{T}|} \sum_{d \in \mathcal{T}} \text{Score}_{\text{method}}(d)
\end{equation}
where the subscript $\text{method}$ can be substituted with $\text{best}$, $\text{avg}$, or $\text{worst}$ to obtain the corresponding overall performance score.

\begin{table*}[h]
    \centering
    \renewcommand{\arraystretch}{1}
    \begin{tabular*}{\linewidth}{@{\extracolsep{\fill}}lccccccc}
        \toprule
        \multirow{2}{*}{\textbf{Model}} &
        \multicolumn{3}{c}{\textbf{PDK}} &
        \multicolumn{2}{c}{\textbf{Synthesis Opt}} &
        \multicolumn{2}{c}{\textbf{Synthesis Tool}} \\
        \cmidrule(lr){2-4}\cmidrule(lr){5-6}\cmidrule(lr){7-8}
        & \textbf{Sky130} & \textbf{gf180} & \textbf{ihp\_sg13g2}
        & \textbf{Timing} & \textbf{Area}
        & \textbf{Yosys} & \textbf{DC} \\
        \midrule
        RTLCoder    & 2.33 & 2.11 & 2.30 & 2.72 & 2.66 & 2.43 & 0.97 \\
        CodeV       & 2.52 & 2.21 & 2.50 & 2.91 & 3.00 & 2.53 & 1.78 \\
        Origen      & 1.46 & 1.37 & 1.41 & 1.52 & 1.59 & 1.45 & 0.97 \\
        GPT-4       & 0.97 & 0.97 & 0.97 & 0.99 & 0.97 & 1.02 & 0.99 \\
        GPT-o1      & 1.03 & 1.04 & 1.01 & 1.00 & 1.07 & 1.06 & 1.00 \\
        ChipSeek & 0.77 & 0.78 & 0.81 & 0.82 & 0.78 & 0.76 & 0.81 \\
        \bottomrule
    \end{tabular*}
    \caption{Results under different PDKs and synthesis configurations. Values are normalized to the benchmark baseline.}
    \label{tab:pdk_synth}
\end{table*}

\section{Detailed Results and More Ablation Study}

To comprehensively evaluate the effectiveness of our proposed training pipeline, we conduct a series of ablation studies. We begin by analyzing the impact of each training stage on functional correctness using the VerilogEval benchmark. The results, presented in Table \ref{ablation_study_verilogeval}, demonstrate the contribution of Supervised Fine-Tuning (SFT) and the subsequent reinforcement learning stage (CDPO). Across all four base models, the application of SFT provides an improvement in `pass@k` metrics. The introduction of the CDPO stage further elevates performance, leading to significant gains in functional correctness.
\newcolumntype{Y}{>{\centering\arraybackslash}X}

We then extend our ablation study to assess the impact of the training stages on PPA metrics. As shown in Table \ref{ablation_rtllm}, while the SFT stage continues to improve functional metrics (`Syntax' and `pass@5'), it sometimes leads to a degradation in performance metrics. This observation underscores a critical challenge: optimizing for functionality does not guarantee optimization for PPA. However, the subsequent CDPO stage effectively addresses this issue. The results clearly indicate that CDPO not only enhances functional correctness further but also significantly improves all PPA metrics, bringing most values below the 1.0 baseline. This demonstrates the crucial role of our reinforcement learning approach in optimizing for real-world hardware design constraints.

Table~\ref{tab:ablation_main} presents an ablation study of ChipSeek built on the \textbf{Qwen2.5-7B} base model. We compare the full system with three variants: \textsc{-AdvAgg} (removing \emph{advantage-level aggregation}, signals are aggregated in reward level), \textsc{-Curriculum} (removing the \emph{curriculum weight schedule}, process rewards are assigned with the constant weights), and \textsc{-PreVec} (removing the \emph{preference vector}, ppa rewards are assigned with the constant weights). All settings use the same training setup, EDA toolchain, and evaluation pipeline.

We report two types of results. \textbf{(i) No-preference evaluation:} \textit{Syntax}, \textit{Pass@5}, \textit{Synthesis}, and \textit{EDAP} are measured on prompts without any PPA preference, reflecting overall correctness and synthesizability. \textbf{(ii) Preference-conditioned evaluation:} \textit{Power}, \textit{Area}, and \textit{Delay} (and their composites \textit{ADP/EDP/EDAP}) are measured under the corresponding preference prompts (e.g., \textit{Power} under \texttt{power} prompts). All PPA metrics are normalized to the reference, where lower is better.

The full ChipSeek achieves the best overall performance. Removing \textbf{AdvAgg} reduces both correctness and PPA quality (e.g., \textit{Pass@5} 84.1$\rightarrow$78.1 and \textit{EDAP} 0.88$\rightarrow$0.92), indicating that advantage-space aggregation helps stabilize multi-objective optimization. Removing \textbf{Curriculum} causes the largest drop in end-to-end success (\textit{Pass@5} 84.1$\rightarrow$72.5 and \textit{Synthesis} 70.91$\rightarrow$63.2) and also worsens \textit{EDAP} (0.88$\rightarrow$0.93), showing the curriculum is important for transitioning from process compliance to PPA optimization. Finally, removing \textbf{PreVec} keeps syntax-level capability similar but weakens preference-following and PPA outcomes (e.g., \textit{Power} 0.93$\rightarrow$0.95 and \textit{EDAP} 0.88$\rightarrow$0.93), confirming that explicit preference conditioning is necessary for controllable PPA optimization.

Next, we present a detailed case-by-case EDAP comparison against the original benchmark designs in Table \ref{ppa_compare}. This table provides raw PPA metrics for a diverse set of hardware modules. The results marked in bold indicate instances where ChipSeek generated a design with superior PPA performance compared to the benchmark. As the data shows, our model successfully optimizes a wide variety of designs, often achieving significant improvements in one or more PPA metrics.

\begin{figure*}[!t] 
    \centering
    \includegraphics[width=\linewidth]{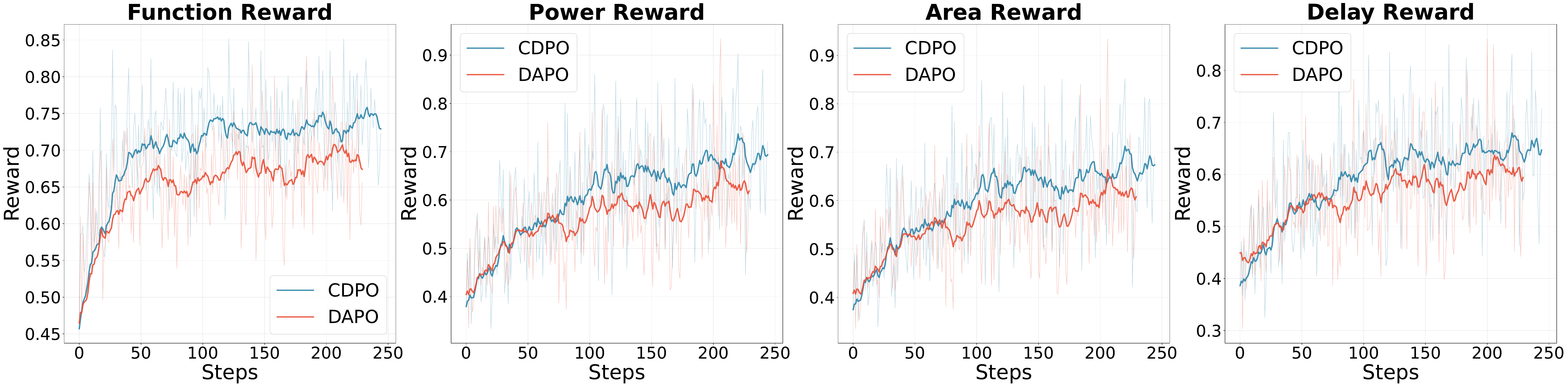}
    \caption{Core reward trajectories comparison between the proposed CDPO and DAPO. CDPO demonstrates superior convergence stability and achieves higher final reward values across all optimization objectives compared to the DAPO baseline. Training steps are mismatched because of the filtering mechanism that omits the equal-reward samples in the training.
}
    \label{fig:cdpo_dapo_rewards}
\end{figure*}

\begin{figure*}[!t] 
    \centering
    \includegraphics[width=0.75\linewidth]{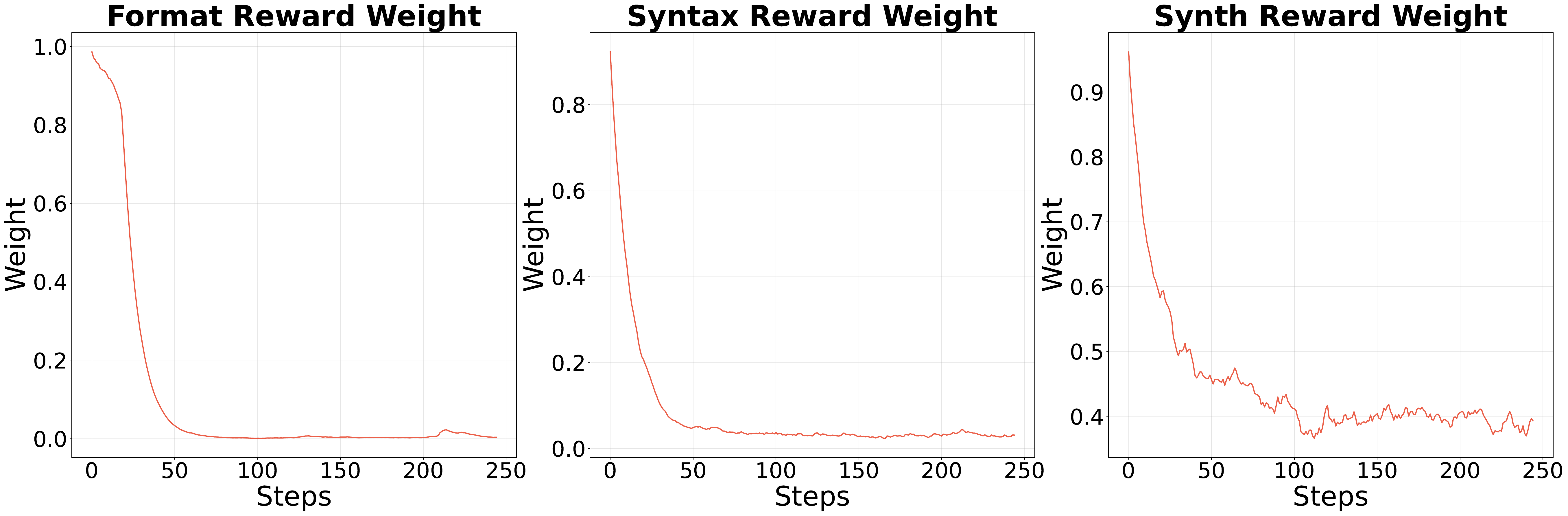}
    \caption{Visualization of the curriculum weight schedule in CDPO.
}
    \label{fig:curriculum_weights}
\end{figure*}

\section{Sensitivity Study}
\label{sec:sensitivity}

We conduct a sensitivity study to assess whether our PPA gains generalize beyond a single toolchain setting. Concretely, we evaluate \textbf{average normalized EDAP} on the RTLLM benchmark while varying three production-level factors: \textit{(i)} standard-cell libraries (PDKs), \textit{(ii)} synthesis tools, and \textit{(iii)} synthesis optimization strategies. This analysis probes the robustness of ChipSeek to common sources of variability in EDA flows.

Starting from the same set of RTLLM tasks, we keep the evaluation protocol fixed and only change one component of the synthesis flow at a time.
We consider three representative PDK libraries (\texttt{Sky130}, \texttt{gf180}, \texttt{ihp\_sg13g2}), two synthesis tools (open-source \texttt{Yosys} and commercial \texttt{Design Compiler (DC)}), and two typical optimization targets (\textit{timing}-driven vs.\ \textit{area}-driven synthesis).
Following prior RTLLM evaluation, EDAP values are normalized to the benchmark baseline, where smaller values indicate better PPA efficiency.

Table~\ref{tab:pdk_synth} reports the sensitivity results. Across all configurations, ChipSeek consistently achieves the lowest normalized EDAP among all methods, while several baselines exhibit substantial degradation under tool or PDK shifts (e.g., larger variance across libraries and more pronounced sensitivity to timing/area targets). In contrast, ChipSeek maintains stable improvements across \textit{all} tested PDKs, synthesis tools, and optimization strategies, indicating that the model is not overfitting to a single library or a particular synthesis recipe.

These results suggest that ChipSeek learns a \emph{robust Verilog generation style} that transfers across realistic EDA settings. The consistent EDAP gains under different PDK libraries, synthesis tools, and optimization objectives provide evidence that our improvements stem from better structural and micro-architectural coding patterns, rather than artifacts of a specific toolchain configuration.

\section{Training Analysis}
\label{sec:training_analysis}

Figure~\ref{fig:cdpo_dapo_rewards} compares the training trajectories of core rewards between \textsc{CDPO} and the \textsc{DAPO} baseline, including the function reward and the three PPA rewards. Across all objectives, \textsc{CDPO} exhibits more stable convergence and reaches consistently higher final reward values. This indicates that \textsc{CDPO} is able to allocate optimization capacity to the truly performance-critical signals, rather than overfitting to shallow constraints.

A key factor behind this behavior is the curriculum weight schedule in \textsc{CDPO}. As visualized in Figure~\ref{fig:curriculum_weights}, the weights of process rewards (format, syntax, and synthesis) decay rapidly as training progresses, effectively shifting the learning emphasis from \emph{easy-to-satisfy} constraints to \emph{hard-to-optimize} core objectives. In the early stage, emphasizing process rewards helps the policy quickly learn valid Verilog formatting, syntactic correctness, and basic synthesizability, providing a reliable foundation for downstream EDA evaluation. Once these prerequisites become consistently achievable, their diminishing weights prevent them from dominating the gradient and allow the policy to focus on improving functional correctness and PPA metrics in later stages. This ``easy-to-hard'' curriculum makes the reinforcement learning process more progressive and better aligned with the ultimate goal of producing high-quality hardware designs.

In contrast, \textsc{DAPO} lacks an explicit curriculum mechanism to down-weight process-level feedback. As a result, even in later training, relatively simple rewards (e.g., format or syntax compliance) can still contribute non-trivially to the overall optimization signal, introducing interference and reducing the effective learning signal for core objectives. This can lead the policy to become overly satisfied with meeting superficial constraints, slowing down further specialization toward functionally correct and PPA-optimized Verilog generation. Overall, Figures~\ref{fig:cdpo_dapo_rewards} and~\ref{fig:curriculum_weights} together support that curriculum-guided reweighting is essential for \textsc{CDPO} to achieve stronger and more stable optimization on the most challenging objectives.

\section{Scalability Analysis}

\begin{table}
    \centering
    \small
    \begin{tabular}{lcc}
        \toprule
        \textbf{Model} & \textbf{Pass@5} & \textbf{Avg Normalized EDAP} \\
        \midrule
        RTLCoder   & 5.6\%  & 1.34 \\
        CodeV      & 8.9\%  & 1.27 \\
        Origen     & 6.7\%  & 1.03 \\
        GPT-4      & 27.8\% & 0.99 \\
        GPT-o1     & 34.4\% & 1.01 \\
        \textbf{ChipSeek} & \textbf{53.3\%} & \textbf{0.86} \\
        \bottomrule
    \end{tabular}
    \caption{Scalability results on large-scale designs from AutoSilicon (e.g., CPU/NPU blocks, avg. $\sim$560 LoC).}
    \label{tab:autosilicon_scalability}
\end{table}

To evaluate scalability on more complex RTL generation tasks, we conduct experiments on large-scale designs from AutoSilicon \cite{autosilicon} (e.g., CPU and NPU blocks, \emph{avg.} $\sim$560 LoC). As shown in Table~\ref{tab:autosilicon_scalability}, ChipSeek achieves a substantially higher Pass@5 (53.3\%) than the strongest baseline (34.4\%), indicating improved functional reliability as program length and structural complexity increase. Meanwhile, ChipSeek also attains the best Avg. Normalized EDAP (0.86), outperforming prior methods and large general-purpose LMs (e.g., GPT-4 at 0.99), which suggests that our EDA-integrated optimization remains effective in steering generation toward better PPA even under a much larger design space. Overall, these results confirm that ChipSeek scales favorably to production-level RTL designs, improving both correctness and hardware quality on long and complex modules.

\begin{figure*}[t]

\begin{tcolorbox}[
    colback=white,
    colframe=black,
    fonttitle=\bfseries,
    title=Barrel Shifter Design Description
]

Solve the following coding problem using the programming language Verilog. Output the code between \verb|```verilog| and \verb|```|.

Please act as a professional verilog designer.

Module name: barrel\_shifter

Function:

A barrel shifter for rotating bits efficiently. This 8-bit barrel shifter takes an 8-bit input and shifts or rotates the bits based on a 3-bit control signal.

Input ports:

    in [7:0]: 8-bit input to be shifted.
    
    ctrl [2:0]: 3-bit control signal that determines how many positions to shift. The control signal specifies the amount of shift, where each bit indicates a shift by 1, 2, or 4 positions.

Output ports:

    out [7:0]: 8-bit shifted output.

Implementation:

    The barrel shifter shifts the input `in` based on the value of the `ctrl` signal. The shifting happens in stages, where each stage shifts by 4, 2, or 1 position, as controlled by `ctrl[2]`, `ctrl[1]`, and `ctrl[0]`, respectively.
    
    The shifts are achieved using a series of 2-to-1 multiplexers (`mux2X1`) that select between the shifted input and the original input. Each stage of multiplexers performs the shift, and the result is passed to the next stage for further shifting if required.
    
    - When `ctrl[2]` is high, the input is shifted by 4 positions.
    
    - When `ctrl[1]` is high, the intermediate result is shifted by 2 positions.
    
    - When `ctrl[0]` is high, the intermediate result is shifted by 1 position.
    
    The `mux2X1` submodule is used to select between two input values based on the `sel` signal, determining which bit is used in the shifted output at each stage.

Please generate the Verilog code optimized for area.
\end{tcolorbox}

\caption{Barrel Shifter Design Description}
\end{figure*}

\begin{figure*}[t]
\begin{tcolorbox}[
    colback=white,
    colframe=black,
    fonttitle=\bfseries,
    title=Golden Solution
]
\begin{lstlisting}[style=mystyle]
module barrel_shifter (
    input  [7:0] in,
    input  [2:0] ctrl,
    output [7:0] out
);

wire [7:0] x;
wire [7:0] y;

// 4bit shift right
mux2X1  ins_17 (.in0(in[7]), .in1(1'b0), .sel(ctrl[2]), .out(x[7]));
mux2X1  ins_16 (.in0(in[6]), .in1(1'b0), .sel(ctrl[2]), .out(x[6]));
mux2X1  ins_15 (.in0(in[5]), .in1(1'b0), .sel(ctrl[2]), .out(x[5]));
mux2X1  ins_14 (.in0(in[4]), .in1(1'b0), .sel(ctrl[2]), .out(x[4]));
mux2X1  ins_13 (.in0(in[3]), .in1(in[7]), .sel(ctrl[2]), .out(x[3]));
mux2X1  ins_12 (.in0(in[2]), .in1(in[6]), .sel(ctrl[2]), .out(x[2]));
mux2X1  ins_11 (.in0(in[1]), .in1(in[5]), .sel(ctrl[2]), .out(x[1]));
mux2X1  ins_10 (.in0(in[0]), .in1(in[4]), .sel(ctrl[2]), .out(x[0]));

// 2 bit shift right
mux2X1  ins_27 (.in0(x[7]), .in1(1'b0), .sel(ctrl[1]), .out(y[7]));
mux2X1  ins_26 (.in0(x[6]), .in1(1'b0), .sel(ctrl[1]), .out(y[6]));
mux2X1  ins_25 (.in0(x[5]), .in1(x[7]), .sel(ctrl[1]), .out(y[5]));
mux2X1  ins_24 (.in0(x[4]), .in1(x[6]), .sel(ctrl[1]), .out(y[4]));
mux2X1  ins_23 (.in0(x[3]), .in1(x[5]), .sel(ctrl[1]), .out(y[3]));
mux2X1  ins_22 (.in0(x[2]), .in1(x[4]), .sel(ctrl[1]), .out(y[2]));
mux2X1  ins_21 (.in0(x[1]), .in1(x[3]), .sel(ctrl[1]), .out(y[1]));
mux2X1  ins_20 (.in0(x[0]), .in1(x[2]), .sel(ctrl[1]), .out(y[0]));

// 1 bit shift right
mux2X1  ins_07 (.in0(y[7]), .in1(1'b0), .sel(ctrl[0]), .out(out[7]));
mux2X1  ins_06 (.in0(y[6]), .in1(y[7]), .sel(ctrl[0]), .out(out[6]));
mux2X1  ins_05 (.in0(y[5]), .in1(y[6]), .sel(ctrl[0]), .out(out[5]));
mux2X1  ins_04 (.in0(y[4]), .in1(y[5]), .sel(ctrl[0]), .out(out[4]));
mux2X1  ins_03 (.in0(y[3]), .in1(y[4]), .sel(ctrl[0]), .out(out[3]));
mux2X1  ins_02 (.in0(y[2]), .in1(y[3]), .sel(ctrl[0]), .out(out[2]));
mux2X1  ins_01 (.in0(y[1]), .in1(y[2]), .sel(ctrl[0]), .out(out[1]));
mux2X1  ins_00 (.in0(y[0]), .in1(y[1]), .sel(ctrl[0]), .out(out[0]));

endmodule

module mux2X1 (
    input in0,
    input in1,
    input sel,
    output out
);

assign out = (sel) ? in1 : in0;

endmodule
\end{lstlisting}
\end{tcolorbox}
\caption{Barrel Shifter Golden Solution}
\label{fig:barrel_shifter_gold}
\end{figure*}

\begin{figure*}[t]
\begin{tcolorbox}[
    colback=white,
    colframe=black,
    fonttitle=\bfseries,
    title=ChipSeek Solution
]
\texttt{\textless think\textgreater}

The implementation prioritizes hardware efficiency by using a logarithmic approach with three cascaded shift stages. Each stage conditionally applies a power-of-two shift based on the corresponding control bit while minimizing area through direct hardware translation without complex multiplexers. This approach eliminates area-intensive components like AND gates and optimizes for area usage.

\texttt{\textless /think\textgreater}

\texttt{\textless answer\textgreater}
\begin{lstlisting}[style=mystyle]
module barrel_shifter (
  input [7:0] in,
  input [2:0] ctrl,
  output [7:0] out
);

// Intermediate wires for shift stages
wire [7:0] shift_4, shift_2, shift_1;

// 4-bit shift stage (controlled by ctrl[2])
assign shift_4 = ctrl[2] ? (in >> 4) : in;

// 2-bit shift stage (controlled by ctrl[1])
assign shift_2 = ctrl[1] ? (shift_4 >> 2) : shift_4;

// 1-bit shift stage (controlled by ctrl[0])
assign shift_1 = ctrl[0] ? (shift_2 >> 1) : shift_2;

assign out = shift_1;

endmodule
\end{lstlisting}
\texttt{\textless /answer\textgreater}
\end{tcolorbox}
\caption{Barrel Shifter ChipSeek Solution}
\label{fig:barrel_shifter_ChipSeek}
\end{figure*}

\section{Detailed Case Study}

In this section, we illustrate the model’s reasoning trajectory and show how it improves the design in both functional correctness and PPA.

A barrel shifter is a key building block in high-performance systems. Unlike a conventional shifter, it can perform multi-bit shifts within a single clock cycle. A common textbook implementation uses a hierarchy of multiplexers, where each stage shifts by a power-of-two offset (e.g., 1, 2, or 4 bits), as shown in the golden solution in Figure \ref{fig:barrel_shifter_gold}. In this task, the human-provided specification highlights a mux2X1-based construction as a clear structural reference.

Importantly, while preserving the required I/O behavior and functional semantics, our model identifies a more synthesis-friendly realization: it expresses the same shifting logic using a concise behavioral description rather than explicitly instantiating a MUX hierarchy. This formulation exposes the underlying logic more directly to the synthesis engine, enabling the backend EDA toolchain to explore a larger optimization space and select the most efficient gate-level mapping under the target library and constraints.

As a result, compared to a literal mux2X1-instantiated implementation, the model’s final design achieves a 13.3\% reduction in area and a 63.9\% reduction in code size, as shown in Figure \ref{fig:barrel_shifter_ChipSeek}. This case study highlights the model’s ability to satisfy human intent (functional requirements and design constraints) while producing implementations that better align with downstream synthesis optimization, ultimately improving hardware metrics.

\section{Prompt Templates used in Automatic Data Augmentation}

This section outlines the structured prompt engineering methodology used in the automatic data augmentation. The approach leverages a series of specialized prompt templates to guide the language model toward specific objectives, including comprehensive function verification and design optimization across various metrics such as delay, area, and power.

\subsection{Testbench Generation Prompt}
The generation of testbench data for function verification is initiated using the following prompt template. This template instructs the model to act as a Verilog testbench engineer, create a sufficient number of test cases based on design complexity, report the outcome, and terminate the simulation correctly.

\begin{tcolorbox}[
    colback=white,
    colframe=black,
    fonttitle=\bfseries,
    title=Testbench Generation Prompt Template
]
You are a professional Verilog testbench engineer. You should use at least 10 test cases to verify the design. The number of test cases should be based on the difficulty of the design. If the design passes all the test cases, please display ``Design passed". If the design does not pass all the test cases, please display ``Design failed with \textless error count\textgreater \space errors out of \textless total count\textgreater \space test cases". Remember to use the \$finish instruction in the end.

Please generate the testbench code for the Verilog code below:

Verilog description: \{instruction\}

Verilog code: \{output\}

Please generate the testbench code below:
\end{tcolorbox}

\subsection{System and Code Formatting Prompts}
To enhance the quality and utility of the model's output, two key prompts are prepended to the main task.

\subsubsection{Thinking System Prompt}
To encourage a detailed and well-reasoned generation process, a "thinking" system prompt is employed. This prompt instructs the model to first articulate its reasoning as an internal monologue before providing the final answer, ensuring a more transparent and logical workflow.

\begin{tcolorbox}[
    colback=white,
    colframe=black,
    fonttitle=\bfseries,
    title=Thinking System Prompt
]
You are a helpful AI Assistant that provides well-reasoned and detailed responses. You first think about the reasoning process as an internal monologue and then provide the user with the answer. Respond in the following format: \\
\textless think\textgreater\textbackslash n...\textbackslash n\textless/think\textgreater \\
\textless answer\textgreater\textbackslash n...\textbackslash n\textless/answer\textgreater
\end{tcolorbox}

\subsubsection{Code Guiding Prompt}
To ensure the generated Verilog code can be programmatically extracted and parsed, the following code-guiding prompt is used. It specifies a clear demarcation for the code block.

\begin{tcolorbox}[
    colback=white,
    colframe=black,
    fonttitle=\bfseries,
    title=Code Guiding Prompt
]
Solve the following coding problem using the programming language Verilog. Output the code between \verb|```verilog| and \verb|```|.
\end{tcolorbox}

\subsection{Design Optimization Prompts}
For hardware design generation, the data augmentation system incorporates prompts targeting specific optimization goals. Each data sample is augmented with one prompt randomly selected from a pool corresponding to one of five optimization priorities.

\subsubsection{Timing/Delay Priority}
When \textbf{timing performance} is the primary optimization objective, one of the following prompts is used:
\begin{tcolorbox}[
    colback=white,
    colframe=black,
    fonttitle=\bfseries,
    title=Prompt Templates with Delay as the Design Priority
]
1. Please generate the Verilog code optimized for timing performance.

2. Focus on minimizing delay and maximizing speed in your Verilog implementation.

3. Optimize your Verilog design for high-speed operation.

4. Please implement the Verilog code with timing performance as the primary goal.
\end{tcolorbox}

\subsubsection{Area Priority}
When \textbf{area efficiency} is the priority, one of the following prompts is selected:
\begin{tcolorbox}[
    colback=white,
    colframe=black,
    fonttitle=\bfseries,
    title=Prompt Templates with Area as the Design Priority
]
1. Please generate the Verilog code optimized for area.

2. Focus on minimizing the hardware area in your Verilog implementation.

3. Optimize your Verilog design for minimal silicon area usage.

4. Please implement the Verilog code with area efficiency as the primary goal.
\end{tcolorbox}

\subsubsection{Power Priority}
For \textbf{low-power design}, the model is guided by one of the following prompts:
\begin{tcolorbox}[
    colback=white,
    colframe=black,
    fonttitle=\bfseries,
    title=Prompt Templates with Power as the Design Priority
]
1. Please generate the Verilog code optimized for power consumption.

2. Focus on minimizing power consumption in your Verilog implementation.

3. Optimize your Verilog design for low power operation.

4. Please implement the Verilog code with power efficiency as the primary goal.
\end{tcolorbox}

\subsubsection{Delay and Power Priority}
To address trade-offs between \textbf{delay and power}, the following prompts are utilized:
\begin{tcolorbox}[
    colback=white,
    colframe=black,
    fonttitle=\bfseries,
    title=Prompt Templates with Delay and Power as the Design Priorities
]
1. Please generate the Verilog code optimized for both delay and power efficiency.

2. Focus on balancing delay and power consumption in your Verilog implementation.

3. Optimize your Verilog design for efficient power usage and high performance.
\end{tcolorbox}

\subsubsection{Delay and Area Priority}
Similarly, for balancing \textbf{delay and area}, the following prompts are used:
\begin{tcolorbox}[
    colback=white,
    colframe=black,
    fonttitle=\bfseries,
    title=Prompt Templates with Delay and Area as the Design Priority
]
1. Please generate the Verilog code optimized for both area efficiency and timing performance.

2. Focus on balancing area usage and speed in your Verilog implementation.

3. Optimize your Verilog design for efficient area usage and high performance.
\end{tcolorbox}

\subsection{Final Prompt Composition}
The final prompt submitted to the model is constructed by systematically concatenating the aforementioned components. Let the primary components be defined as:
\begin{itemize}
    \item $P_{\text{guide}}$: The \textit{Code Guiding Prompt}.
    \item $D_{\text{desc}}$: The \textit{Verilog Design Description} (i.e., the instruction).
    \item $P_{\text{opt}}$: The \textit{Design Priority Prompt}, selected from one of the optimization pools.
    \item $S_{\text{sys}}$: The \textit{Thinking System Prompt}.
\end{itemize}

First, the user-facing prompt, $P_{\text{user}}$, is constructed by concatenating the guiding prompt, the design description, and the optimization prompt. Let $\oplus$ denote the string concatenation operation.
$$
P_{\text{user}} = P_{\text{guide}} \oplus D_{\text{desc}} \oplus P_{\text{opt}}
$$

Next, the final prompt, $P_{\text{final}}$, is assembled by prepending the system prompt and its associated keywords to the user-facing prompt.
$$
P_{\text{final}} = \text{``system\_prompt:"} \oplus S_{\text{sys}} \oplus \text{``user\_prompt:"} \oplus P_{\text{user}}
$$
This structured composition ensures that the model receives clear, multi-faceted instructions tailored to the specific generation task.

\begin{table*}[!h]
\centering
\renewcommand{\arraystretch}{1.15}
\small
\begin{tabularx}{\textwidth}{|l|Y|Y|}
\hline
\multirow{2}{*}{Name} & Original Benchmark & \textbf{ChipSeek} \\ & $(\mathrm{ns}, \mu\mathrm{m}^2, \mathrm{W})$ & $(\mathrm{ns}, \mu\mathrm{m}^2, \mathrm{W})$ \\
\hline
asyn\_fifo & \textbf{0.72/1397.032/7.67e-05} & N/A \\
\hline
LFSR & 0.14/25.004/2.45e-06 & 0.14/25.004/2.45e-06 \\
\hline
right\_shifter & 0.08/36.176/4.32e-06 & 0.08/36.176/4.32e-06 \\
\hline
barrel\_shifter & 0.17/44.688/1.41e-05 & \textbf{0.17/39.368/1.41e-05} \\
\hline
LIFObuffer & 0.38/226.1/0.00027 & \textbf{0.36/216.79/0.000137} \\
\hline
RAM & 0.25/635.74/5.56e-05 & \textbf{0.19/475.076/3.92e-05} \\
\hline
ROM & 0.14/6.65/8.85e-07 & 0.14/6.65/8.85e-07 \\
\hline
alu & 1.92/1573.39/0.000751 & \textbf{1.75/1286.908/0.000433} \\
\hline
pe & 1.27/3651.382/0.000224 & 1.27/3651.382/0.000224 \\
\hline
instr\_reg & 0.14/117.04/1.03e-05 & 0.14/117.04/1.03e-05 \\
\hline
signal\_generator & 0.37/93.1/7.54e-06 & \textbf{0.38/74.214/6.08e-06} \\
\hline
square\_wave & 0.41/100.282/8.39e-06 & 0.41/100.282/8.39e-06 \\
\hline
calendar & 0.44/164.92/1.44e-05 & 0.44/164.92/1.44e-05 \\
\hline
parallel2serial & 0.2/48.678/4.5e-06 & \textbf{0.19/47.082/4.36e-06} \\
\hline
pulse\_detect & 0.18/17.556/1.52e-06 & \textbf{0.17/16.226/1.47e-06} \\
\hline
serial2parallel & 0.4/156.142/1.4e-05 & \textbf{0.28/157.738/1.39e-05} \\
\hline
width\_8to16 & 0.24/186.732/1.67e-05 & \textbf{0.21/173.698/1.62e-05} \\
\hline
traffic\_light & 0.37/149.758/1.31e-05 & \textbf{0.36/148.162/1.25e-05} \\
\hline
edge\_detect & 0.12/18.354/1.79e-06 & \textbf{0.1/17.822/1.75e-06} \\
\hline
freq\_divbyfrac & \textbf{0.2/48.678/4.67e-06} & N/A \\
\hline
freq\_divbyeven & \textbf{0.25/40.166/3.63e-06} & N/A \\
\hline
freq\_divbyodd & 5.17/59.052/6.63e-06 & \textbf{7.82/82.642/1.33e-6} \\
\hline
sequence\_detector & 0.15/36.442/3.35e-06 & \textbf{0.19/25.27/2.27e-06} \\
\hline
ring\_counter & 0.1/46.816/4.7e-06 & \textbf{0.1/40.964/4.01e-06} \\
\hline
JC\_counter & 0.1/340.48/3.54e-05 & 0.1/340.48/3.54e-05 \\
\hline
counter\_12 & 0.25/36.176/3.1e-06 & 0.25/36.176/3.1e-06 \\
\hline
up\_down\_counter & 0.7/217.854/1.74e-05 & \textbf{0.67/188.86/1.61e-05} \\
\hline
adder\_bcd & 0.34/46.018/3.85e-05 & 0.34/46.018/3.84e-05 \\
\hline
adder\_pipe\_64bit & 0.75/2534.182/0.000235 & \textbf{0.15/886.964/6.49e-5} \\
\hline
adder\_32bit & 0.76/472.15/0.000325 & \textbf{1.13/191.786/0.00012} \\
\hline
adder\_16bit & 0.84/89.376/6.49e-05 & \textbf{0.84/93.632/4.44e-05} \\
\hline
adder\_8bit & 0.35/51.072/3.14e-05 & \textbf{0.35/46.816/2.22e-05} \\
\hline
fixed\_point\_adder & 1.69/606.214/0.000565 & \textbf{1.42/512.313/0.000482} \\
\hline
fixed\_point\_subtractor & 1.09/477.736/0.000381 & \textbf{0.93/466.8/0.000367} \\
\hline
multi\_pipe\_4bit & 0.34/174.762/1.51e-05 & \textbf{0.1/154.546/1.28e-05} \\
\hline
multi\_pipe\_8bit & \textbf{0.8/874.608/7.52e-05} & N/A \\
\hline
multi\_16bit & 2.03/933.394/7.36e-05 & \textbf{1.97/935.522/7.36e-05} \\
\hline
multi\_8bit & 1.5/483.854/0.00085 & \textbf{0.79/373.996/0.000545} \\
\hline
comparator\_4bit & 0.16/18.886/8.91e-06 & \textbf{0.13/17.29/7.6e-06} \\
\hline
comparator\_3bit & 0.1/11.704/5.26e-06 & \textbf{0.1/11.704/5.25e-06} \\
\hline
radix2\_div & \textbf{0.59/414.162/3.39e-05} & N/A \\
\hline
div\_16bit & 5.18/760.228/0.027 & \textbf{5.57/745.332/0.0234} \\
\hline
accu & \textbf{0.47/150.822/1.23e-05} & 0.46/210.672/1.83e-05 \\
\hline
sub\_64bit & 2.3/404.586/0.000268 & \textbf{2.08/400.862/0.000271} \\
\hline
\end{tabularx}
\caption{Comparison of PPA metrics on RTLLM v2.0 between ChipSeek and benchmark. N/A means generated code are not functionally correct or can't pass the scripts of the EDA tools.}
\label{ppa_compare}
\end{table*}

\end{document}